\documentclass[9pt,sigconf]{acmart}

\usepackage{balance}

\usepackage{enumitem}

\usepackage{multirow}
\DeclareMathOperator{\EX}{\mathbb{E}}

\usepackage{graphicx} 

\usepackage[labelfont=bf]{caption}

\usepackage{color}

\usepackage{tabu}

\usepackage{booktabs}
\usepackage{amsmath}

\usepackage{array}
\usepackage{url} 
\usepackage{algorithmicx}
\usepackage{tabularx,booktabs}
\usepackage{graphicx}
\usepackage[footnotesize]{subfigure}
\makeatletter
\newif\if@restonecol
\makeatother

\usepackage[linesnumbered,ruled,vlined]{algorithm2e}
\usepackage{algpseudocode}
\usepackage{amsmath}
\newcolumntype{C}[1]{>{\centering\let\newline\\\arraybackslash\hspace{0pt}}m{#1}}
\usepackage{tabularx} 
\acmConference[SenSys]{The 15th ACM Conference on Embedded Networked Sensor Systems}{November 6-8, 2017}

\renewcommand\footnotetextcopyrightpermission[1]{} 

\newcommand{\aliasAPP}{\textit{DRLIC}\xspace}
\title{Optimizing Irrigation Efficiency using Deep Reinforcement Learning in the Field}

\author{Xianzhong Ding}
\affiliation{%
  \institution{University of California, Merced}
}
\email{xding5@ucmerced.edu}

\author{Wan Du}
\affiliation{%
  \institution{University of California, Merced}
}
\email{wdu3@ucmerced.edu}

\begin{document}

\begin{abstract}

Agricultural irrigation is a significant contributor to freshwater consumption. However, the current irrigation systems used in the field are not efficient. They rely mainly on soil moisture sensors and the experience of growers, but do not account for future soil moisture loss. Predicting soil moisture loss is challenging because it is influenced by numerous factors, including soil texture, weather conditions, and plant characteristics.

This paper proposes a solution to improve irrigation efficiency, which is called DRLIC. DRLIC is a sophisticated irrigation system that uses deep reinforcement learning (DRL) to optimize its performance. The system employs a neural network, known as the DRL control agent, which learns an optimal control policy that considers both the current soil moisture measurement and the future soil moisture loss. We introduce an irrigation reward function that enables our control agent to learn from previous experiences. However, there may be instances where the output of our DRL control agent is unsafe, such as irrigating too much or too little water. To avoid damaging the health of the plants, we implement a safety mechanism that employs a soil moisture predictor to estimate the performance of each action. If the predicted outcome is deemed unsafe, we perform a relatively-conservative action instead. To demonstrate the real-world application of our approach, we developed an irrigation system that comprises sprinklers, sensing and control nodes, and a wireless network.

We evaluate the performance of DRLIC by deploying it in a testbed consisting of six almond trees. During a 15-day in-field experiment, we compared the water consumption of DRLIC with a widely-used irrigation scheme. Our results indicate that DRLIC outperformed the traditional irrigation method by achieving a water savings of up to 9.52\%.


\end{abstract}





\maketitle


 \begin{figure*}[t]
		\begin{minipage}[t]{0.32\linewidth}
		\centering
		 \includegraphics[width=1.4in,height=1.32in,angle=0]{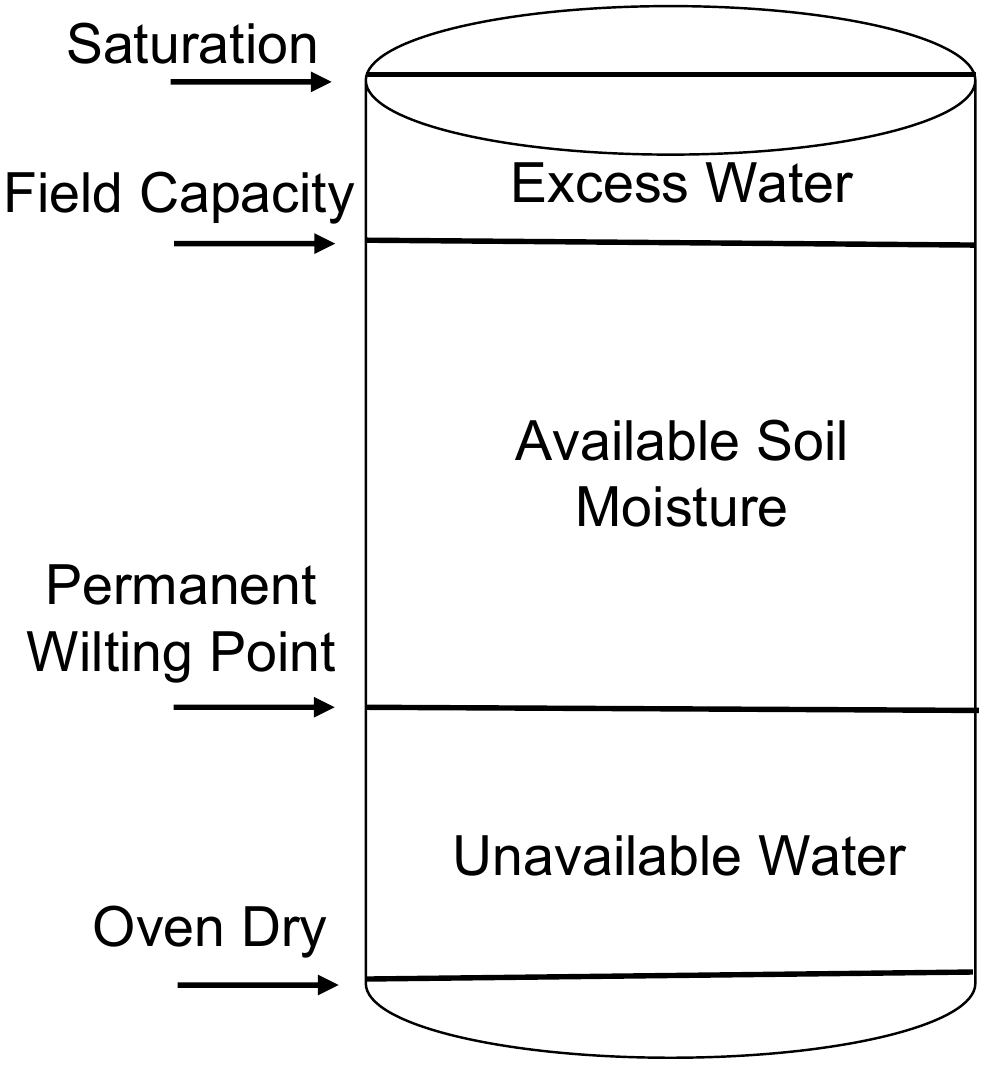}
		\caption{The various levels of the soil water content \cite{fieldcapacity}.}
		\label{soil_level}
	\end{minipage}	
		\hspace{1ex}
		\begin{minipage}[t]{0.32\linewidth}
		\centering
        \includegraphics[width=2.2in,height=1.32in,angle=0]{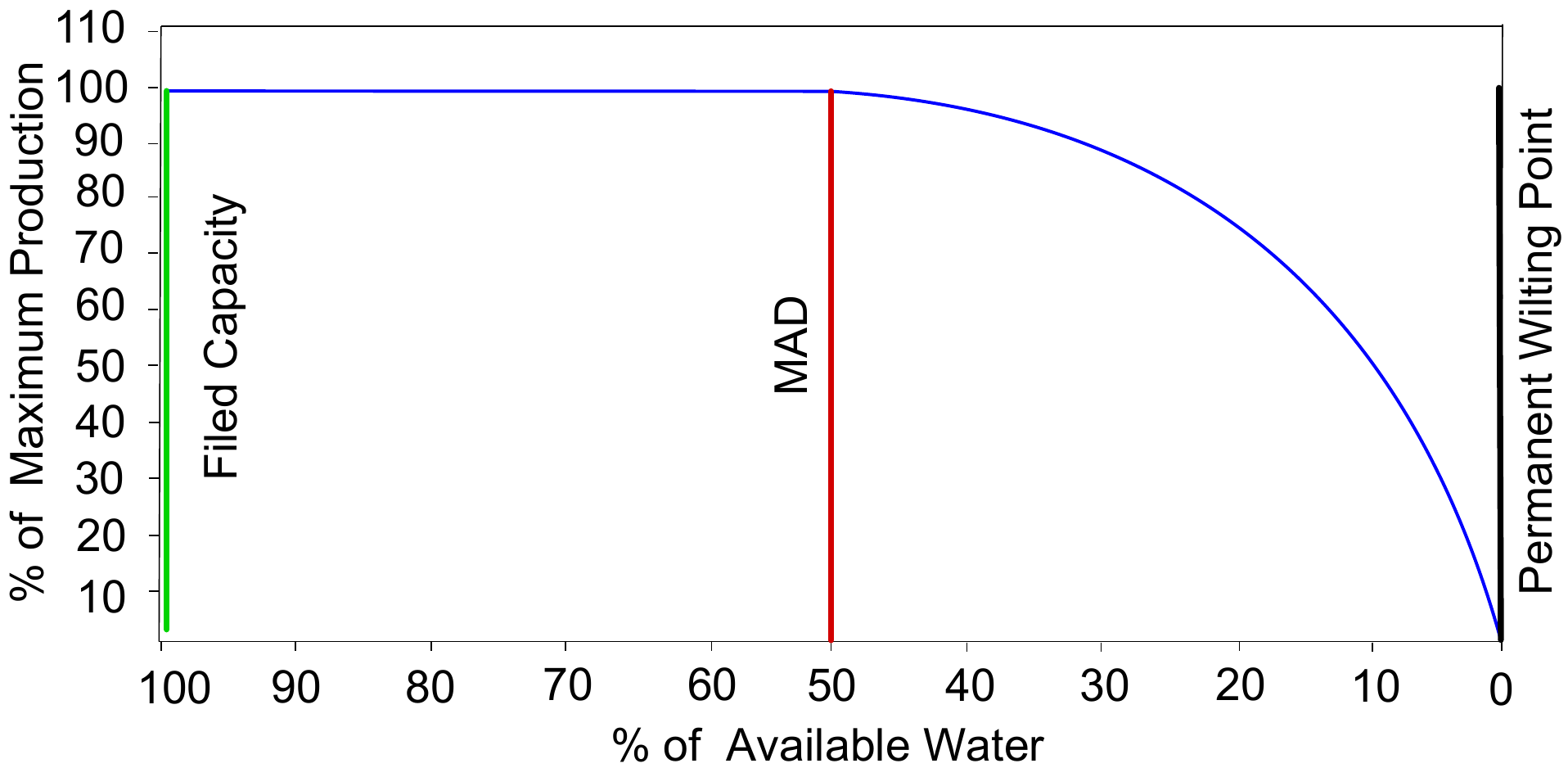}
		\caption{How plant production (growth) is affected by soil water content \cite{peters2013practical}.}
		\label{soil_mad}	 
	\end{minipage}	
	\hspace{1ex}
	\begin{minipage}[t]{0.32\linewidth}
		\centering
		 \includegraphics[width=2.2in,height=1.32in,angle=0]{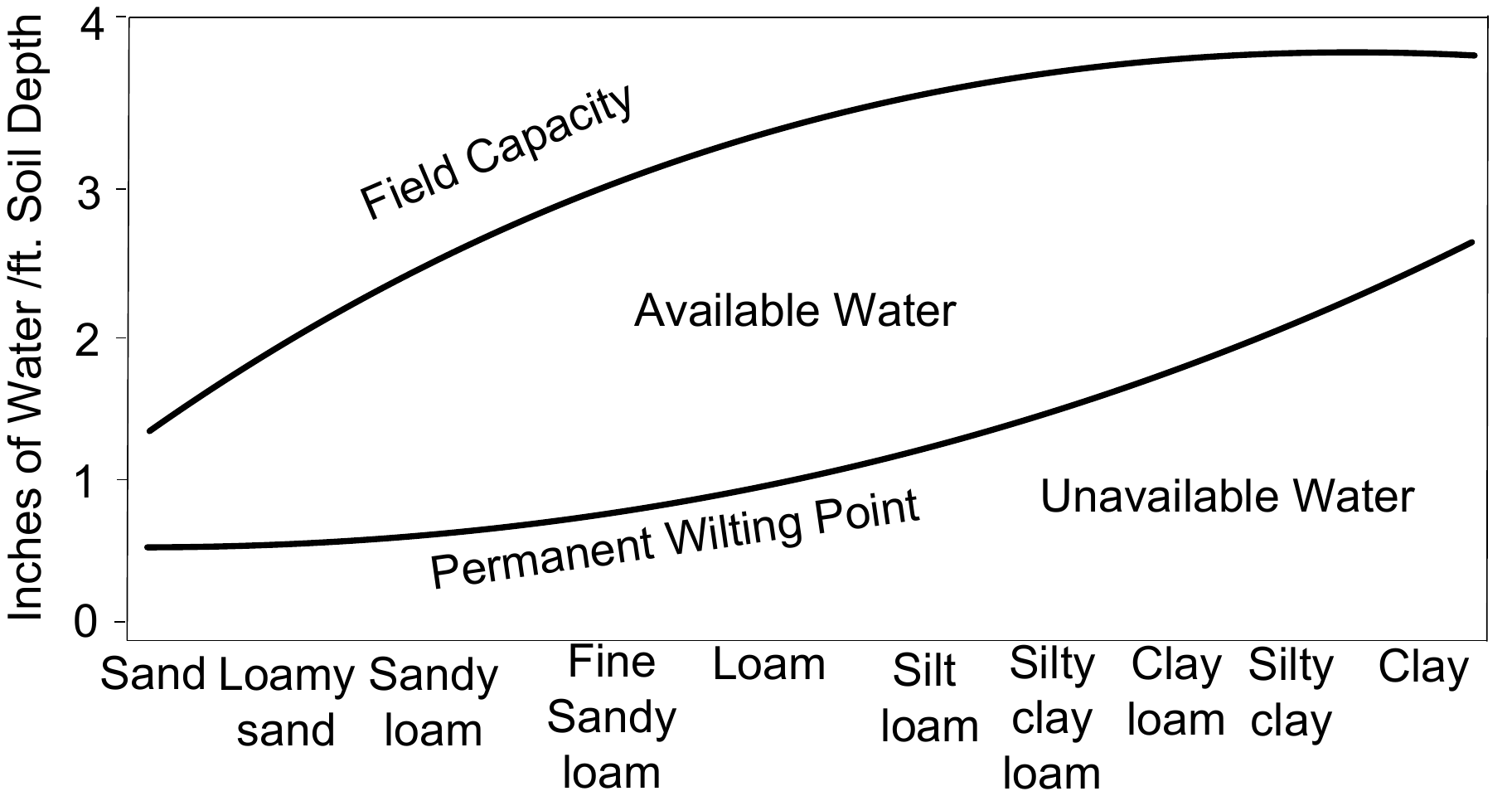}
		\caption{Relationship between available water capacity and soil texture \cite{pwp}.}
		\label{soil_texture}		
	
	\end{minipage}	
\end{figure*}

\section{Introduction}

Agriculture is a major contributor to the consumption of ground and surface water in the United States, with estimates suggesting that it accounts for approximately 80\% of the Nation's water use, and over 90\% in many Western states\footnote{Irrigation and Water Use: https://www.ers.usda.gov/}. Specifically, California's almond acreage in 2019 was estimated at 1,530,000 acres, and almond irrigation alone is estimated to consume roughly 195.26 billion gallons of water annually~\cite{fulton2019water,almond_number}. Given the current drought affecting many Western states, it is critical to improve irrigation efficiency to conserve our limited freshwater reserves. This study focuses on enhancing the irrigation efficiency of almond orchards.


The primary objective of agricultural irrigation is to maintain the health of trees and maximize crop production. Achieving this goal requires maintaining the soil moisture of the trees within a specific range, typically between the Field Capacity (FC) level and the Management Allowable Depletion (MAD) level. If the soil moisture falls below the MAD level, almond trees may experience discoloration or even die. Conversely, if the soil moisture exceeds the FC level, it can lead to reduced oxygen movement in the soil, negatively impacting the tree's ability to absorb water and nutrients. Both FC and MAD levels are dependent on the soil type and plant species. Therefore, to determine the appropriate FC and MAD levels for a specific orchard, it is essential to identify the soil type and refer to a manual that outlines the corresponding FC and MAD levels for that soil type \cite{almond_manual}.


To ensure that the soil moisture remains within the MAD and FC range, the sprinklers must be activated every day or every few days, depending on the soil moisture level. Given the high evaporation rate in California, daily irrigation is recommended by the Almond Board of California \cite{almond_manual} and is used in many existing irrigation systems~\cite{grabow2013water, ding2022drlic}. In most micro-sprinkler irrigation systems, irrigation is performed at night to reduce water loss due to evaporation, which can be as high as 14-19\% during the day \cite{urrego2013relevance}. The irrigation scheduling problem involves determining the appropriate amount of water to be applied to each sprinkler to ensure that the soil moisture remains within the MAD and FC range until the next irrigation cycle. This decision is based on the current soil moisture level and the predicted soil moisture loss for the following day, which is influenced by factors such as soil type, local weather conditions, and plant properties (e.g., root length and leaf number). The objective of irrigation is to provide trees with an appropriate amount of water such that the soil moisture remains above the MAD level until the next irrigation cycle.


Developing optimal irrigation control strategies requires accurate soil moisture loss prediction models. Traditional Model Predictive Control (MPC) methods can be utilized for optimal irrigation control if such a prediction model exists, but the accuracy of the model can have a significant impact on the performance of these methods \cite{delgoda2016irrigation, lozoya2014model}. Obtaining an accurate soil moisture prediction model for an almond orchard is challenging, as soil moisture is influenced by multiple factors, including soil type, topography, ambient temperature, humidity, solar radiation intensity, and plant transpiration \cite{silva2019modelling}. Additionally, customized soil moisture models must be developed for each orchard, limiting the scalability of MPC-based approaches. These two limitations have prevented the use of MPC-based methods in orchards.

The irrigation systems currently used in orchards are ET-based or sensor-based control methods.
Evapotranspiration (ET) is an estimate of moisture lost from soil, subject to weather factors such as wind, temperature, humidity, and solar irradiance. 
All these weather factors are being measured by weather stations.
Local ET value is also publicly available \cite{et_database} and updated every hour.
Based on the ET values since the last irrigation time, ET-based irrigation controllers start the sprinklers to compensate for the soil moisture loss.
However, they do not consider the soil moisture loss of next day before the next irrigation time.
If the soil moisture loss in the last day does not equal the soil moisture loss that will happen in the next day, ET-based irrigation may under-irrigate or over-irrigate.
In addition, a safe margin of water \cite{winkler2018plug} is normally added, making ET-based methods over-irrigate in most cases \cite{grabow2013water, friedman2023crop}.

With accurate soil moisture sensors, irrigation controllers can react directly to the soil moisture level \cite{grabow2013water}. 
The commonly-used controllers are "rule-based", in which a certain amount of water will be supplied once soil moisture deficiency is detected. 
However, parameters for the time and the amount to irrigate are generally tuned by growers by their experience. 
Without predicting how much water will be lost, sensor-based irrigation normally does not systematically take into account future weather information, such as rain and wind in next day.

To solve the limitations of the above existing irrigation schemes, we develop \aliasAPP, a practical Deep Reinforcement Learning (DRL)-based irrigation system, which automatically learns an optimal irrigation control policy by exploring different control actions. 
In \aliasAPP, a control agent observes the \textit{state} of the environment, and chooses an \textit{action} based on a control policy. After applying the action, the environment transits to a next state and the agent receives a \textit{reward} to its \textit{action}. 
The goal of learning is to maximize the expected cumulative discounted reward.
\aliasAPP's control agent uses a neural network to learn its control policy.
The neural network maps "raw" observations to the irrigation decision for the next day. The state includes the weather information (e.g., ET and Precipitation) of today and next day.

To minimize the irrigation water consumption while not impacting the trees' health, we design a reward function that considers three specific situations. If the soil moisture result is higher than the FC level or lower than the MAD level, we will give the control agent a negative reward. If the soil moisture result is within the MAD and FC range, we will give the control agent a positive reward inversely proportional to the water consumption.




Ideally, \aliasAPP's control agent should be trained in a real orchard of almond trees. However, due to the long irrigation interval (one day in our case), the control agent can only explore 365 control actions per year. It will take 384 years to train a converged control agent. 
Therefore, to speed up the training process, we train our control agent in a customized soil-water simulator. 
The simulator is calibrated by the 2-month soil moisture data of six almond trees and can generate sufficient training data for \aliasAPP using 10-year weather data.

Working as an irrigation controller in the field, the control agent may meet some states that it has not seen during training, especially for the control agent trained in a simulated environment. 
In this situation, the control agent may make a poor decision that violates plants' health, i.e., making the soil moisture level lower than the MAD level or higher than the FC level. 
To handle the gap between the simulated environment and the real orchard, we design a safe irrigation mechanism.
If \aliasAPP's control agent outputs an unwise action, instead of executing that action, we use the ET-based method to generate another action. 
We use the soil moisture model of our soil-water simulator to verify whether an action is safe or not. 


To evaluate the performance of \aliasAPP, we build an irrigation testbed with micro-sprinklers currently used in almond orchards. 
Six almond trees are planted in two raise-beds.
Each tree has a sensing and control node, composed of an independently-controllable micro-sprinkler and a soil moisture measurement set (two sensors deployed at different depths in the soil). 
Each node can send its sensing data to our server via IEEE 802.15.4 wireless transmission, and receive irrigation commands from the server.  


Our testbed has been successfully deployed in the field, where we have collected soil moisture data from six sensing and control nodes for over three months. Using two months of this data, we trained our soil moisture simulator, while 0.5 months of data were used to validate its accuracy. Once we trained the control agent of DRLIC, we deployed it in our testbed for 15 days. The results of the experiment showed that DRLIC can reduce water usage by 9.52\% compared to the ET-based control method, without causing any harm to the health of almond trees

We summarize the main contributions of this paper as follows:
\begin{itemize}
\item We design \aliasAPP, an irrigation method that utilizes DRL to save water usage in agriculture. 


\item A set of techniques have been proposed to transform \aliasAPP into a practical irrigation system, including our customized design of DRL states and reward for optimal irrigation, a validated soil moisture simulator for fast DRL training, and a safe irrigation module.



\item We build an irrigation testbed with customized sensing and actuation nodes, and six almond trees. 

\item Extensive experiments in our testbed show the effectiveness of \aliasAPP.
\end{itemize}







\section{Irrigation Problem}
\label{background}



\textbf{Soil Water Content Parameters.} Soil plays a critical role in the water supply for plants, serving as a reservoir for their hydration needs. Typically, up to 35\% of the space in soil can be filled with water. Soil water content is a measure of the amount of water present in the soil, often expressed as a percentage of water by volume (\%) or as inches of water per foot of root (in/ft). Soil moisture sensors are commonly employed to measure the soil water content at a specific location in the soil. For trees with roots that extend several feet, multiple soil moisture sensors may be utilized at various depths along the root structure. The root is typically divided into a defined number of sections, with a soil moisture sensor positioned at the midpoint of each section to enable accurate monitoring of the plant's hydration levels. The soil water content of the tree can be calculated as $V = \sum_{j=1}^{M}\varphi_{j}*d_{j}$, where $M$ is the number of moisture sensors installed at different depths (M is 2 in our experiments);
$\varphi_j$ is the reading measured by the $j$th soil moisture sensor; and $d_j$ is the depth that the $j$th moisture sensor covers. 
If such a set of soil moisture sensors are used to measure the soil water content of a region, they will be deployed under a typical tree that has similar soil water content with most of the trees in the region.


To ensure the health of a plant, it is essential to ensure that its roots have access to a sufficient supply of water. Figure \ref{soil_level} provides an illustration of two critical soil water content levels that are crucial for plant health \cite{fieldcapacity}. Firstly, the Permanent Wilting Point (PWP) represents the minimum threshold of soil water content below which plants are unable to draw sufficient moisture from the soil. Prolonged periods of soil moisture levels below the PWP can lead to plant wilting or death. Secondly, if the soil water content exceeds the Field Capacity (FC) level, there is an excess of water in the soil, which can lead to water wastage and rotting of the roots over time, ultimately compromising the plant's health. Therefore, \textit{the goal of irrigation systems is to maintain soil water content between the PWP level and the FC level}.


\begin{table}[t]
  \renewcommand\arraystretch{0.8}
  \caption{Suggested MAD For Different Crops \cite{peters2013practical}}
\vspace{-0.1in}
  \centering
  \begin{tabular}{c|c||c|c}
    \hline
     \textbf{Crop} & \textbf{MAD (\%)} &\textbf{Crop} & \textbf{MAD (\%)}
\\
    \hline
   Beans   & 40  &Potatoes & 30 \\
    \hline
    Blueberries &  50 &Raspberries & 50 \\
    \hline
   Corn   & 50 & Strawberries &  50\\
    \hline
   Alfalfa   & 55&Sweet Corn & 40\\
    \hline
    Mint &  35 & Tree Fruit &  50\\

    \hline    
  \end{tabular}
  \label{mad_plants}
\end{table}

The primary objective of irrigation for fruit trees like almond is to achieve maximum production. To achieve this goal, it is crucial to maintain soil moisture content above the Management Allowable Depletion (MAD) level instead of the PWP level. As shown in Table \ref{mad_plants}, the MAD level can vary for different types of crops (e.g., 40\% for corn, 50\% for fruit trees). Figure \ref{soil_mad} illustrates the relationship between soil moisture content and almond tree production \cite{peters2013practical}.  From Figure \ref{soil_mad}, it can be observed that the MAD level for almond trees is the median value (50\%) between the Field Capacity (FC) level and the Permanent Wilting Point (PWP) level. Therefore, \textit{almond trees can achieve their maximum production, as long as we maintain the soil water content above the MAD level.}


\textbf{How to Determine these Parameters in an Orchard?} The Available Water holding Capacity (AWC) of the soil is defined as the soil water content range between the FC level and the PWP level. Figure \ref{soil_texture} shows that AWC varies for different soil types \cite{pwp}. The texture, presence, and abundance of rock fragments, as well as the depth and layers of soil, can affect the AWC of the soil. Finer-textured soils, such as loam, have a higher AWC than sandier soils \cite{pwp}. On the other hand, soils with more clay, such as clay loam, have a lower AWC than loamy soils \cite{pwp}. 


The AWC of a tree, $V_{awc}$, can be calculated as  $V_{awc} = \sigma_{awc} * D_{foot}$, where $\sigma_{awc}$ is the soil's AWC and $D_{foot}$ is the tree's root depth in the unit of feet.
The AWC for different soil types, $\sigma_{awc}$, can be found in \cite{pwp}. 

The PWP level for a soil type, $V_{pwp}$, can also be calculated as $V_{pwp} = \varphi_{pwp} * D_{inch}$, where 
$\varphi_{pwp}$ is the soil moisture content at the wilting point of that soil type and $D_{inch}$ are the root depth of the plant in the unit of inches.
$\varphi_{pwp}$ for a specific soil type can be found in \cite{pwp}.

Based on the above two parameter ($V_{awc}$ and $V_{pwp}$), we can also obtain the FC level as $V_{fc} = V_{awc} + V_{pwp}$, and 
the MAD level as $V_{mad} = \alpha * V_{awc} + V_{pwp}$, where $\alpha$ is set to 50\% for almond trees.


\textbf{How to Use these Parameters for Irrigation?}
The main objective of irrigation is to maintain the soil water content of plants between the Field Capacity (FC) level and the Management Allowable Depletion (MAD) level. To achieve this goal, it is crucial to determine the soil’s Available Water holding Capacity (AWC) and the Permanent Wilting Point (PWP) level ($V_{awc}$ and $V_{pwp}$), which depend on the soil type, texture, and layers. Once we identify the soil type, we can calculate these parameters using the methods discussed earlier. However, in large orchards with varying soil types and changing parameters, it is essential to adjust the irrigation system's setting accordingly.


\textbf{How Many Valves to Control in an Orchard?}
Ideally, the sprinkler for each tree should be individually controlled, since the ET of each tree in an orchard varies from 0.12 to 0.20 inches \cite{niu2020estimating}. Moreover, the soil type also varies spatially in an orchard \cite{almond_manual}, e.g., there are 10 soil type differences with soil clay loam accounting for from 45.6\% to  54.7\% and 0 to 8 percent slopes in a 60-acre orchard of California~\footnote{Soil Map: https://casoilresource.lawr.ucdavis.edu/gmap/}.
However, there are around 75-125 almond trees in one acre, it is costly to deploy a soil moisture sensor under each tree. 
Thus, an orchard is normally divided into several irrigation regions based on the similarity of soil texture. 
A valve is used in each irrigation region to control all the sprinklers.
The irrigation problem of a large orchard is to control a number of valves. 
This paper is focused on irrigation scheduling, but not field partitioning. 
A simple way to partition an orchard into several irrigation regions is to survey the soil samples across the orchard using an auger. Growers normally conduct the survey for other purposes too, such as planning the density of trees and fertilizing the trees.

\begin{figure}[t]
\includegraphics[height=2.6in, width=3in]{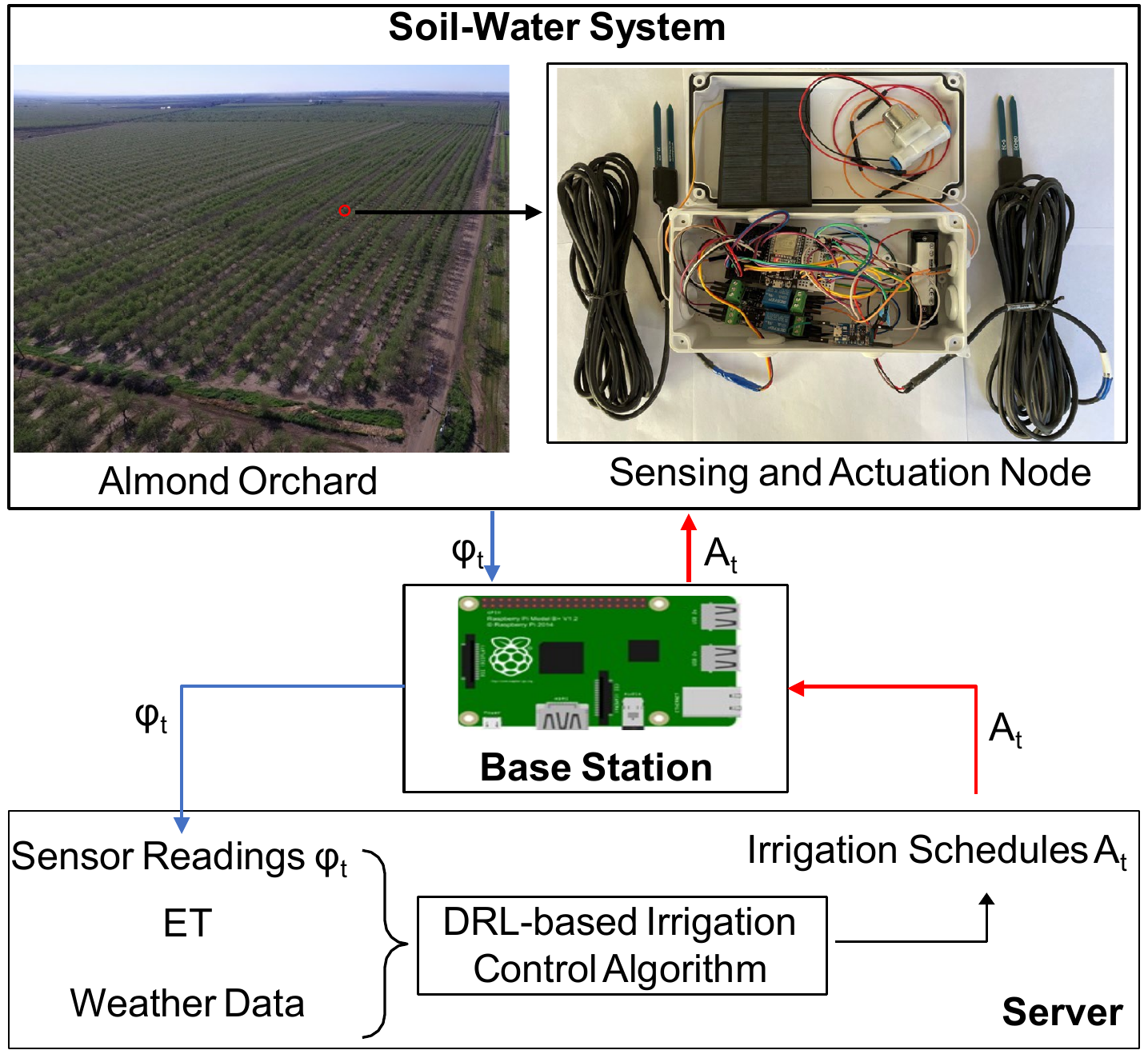}\Description{framework}
  \caption{\aliasAPP System Architecture.}
  \label{framework}
\end{figure}

\section{\aliasAPP System Design}
In this section, we first give an overview of \aliasAPP. We model the irrigation problem as a Markov decision process.
We design a DRL-based irrigation scheme and a safe irrigation module.

\subsection{Overview}
Figure~\ref{framework} shows the system architecture of \aliasAPP, which is composed of two key components, i.e., a wireless network of sensing and actuation sprinkler nodes, and a DRL-based control algorithm. 

For an almond orchard, we install the sensing and actuation node for each irrigation region. One sensing and actuation node is equipped with a set of soil moisture sensors that are deployed at different depths in the soil. 
Sensing data is transmitted to the base station via an IEEE 802.15.4 network. 
The \textit{Base Station} collects the data from \aliasAPP nodes and sends them to a local server using Wi-Fi. 
These sensing data collected from all \aliasAPP nodes creates a “snapshot” of the soil moisture readings $\varphi_{t}$ across the entire orchard. 

On the server, the DRL-based irrigation control agent makes irrigation decisions based on the soil moisture sensors' readings, ET and weather data from local weather stations.
It provides the optimal irrigation schedule for all \aliasAPP nodes. The objective of \aliasAPP is to minimize the total irrigation water consumption while meeting the requirement of almond health.  
The server will send the generated irrigation schedules $A_{t}$ to all \aliasAPP nodes. 
By receiving a command, a node may open its sprinkler by a latching solenoid with two relays. 
The implementation details of the nodes will be introduced in Section \ref{testbed_hardware}.

\subsection{MDP and DRL for Irrigation}
We adopt the daily irrigation scheme, i.e., the irrigation starts at 11 PM every day. 
Each time, the controller decides how long to open each sprinkler to guarantee that the soil water content will be still within the MAD and FC range tomorrow night. 
The future soil water content is determined by the current soil water content, the irrigated water volume, the trees' water absorption, and soil water loss (caused by runoff, percolation and ET). 
Such a sequential decision-making problem can be formulated as a Markov Decision Process (MDP), modeled as \textsl{<S, A, T, R>}, where \begin{itemize}
\item~~ \textsl{S} is a finite set of states, which includes sensed moisture level from orchard and weather data from local station.
\item~~\textsl{A} is a finite set of irrigation actions for all control valves. 
\item~~\textsl{T} is the state transition function defined as $\textsl{T : S} \times \textsl{A}  \rightarrow  \textsl{S}$. 
The soil water content at next time step is determined by current soil water content and the irrigation action.
\item~~\textsl{R} is the reward function defined as $ \textsl{S} \times   \textsl{A}  \rightarrow  \mathbb{R}$, which qualifies the performance of a control action.

\end{itemize} 
Based on the above MDP-based irrigation problem formulation, we will find an optimal control policy  $\pi(s)^{*}  : \textsl{S} \rightarrow  \textsl{A} $, which maximizes the accumulative reward \textsl{R}.  
We cannot apply conventional tools (e.g., dynamic programming) to search for the optimal control policy, because the state transition function is hard to analytically characterize. 
In this paper, we consider an RL-based approach to generating irrigation control algorithms. Unlike previous approaches that use pre-defined rules in heuristic algorithms, our approach will learn an irrigation policy from observations. 


DRL is a data-driven learning method. It has been widely applied in many control applications \cite{zhu2021network, shen2020dmm, liu2020continuous, kumar2021building, ding2019octopus, ding2020mb2c}. 
DRL learns an optimal control policy through interacting with the environment. 
At each time step $t$, the control agent selects an action $A_{t} = a$, given the current state $S_{t} = s $, based on its policy $\pi_{\theta}$.

\begin{equation}
  \begin{aligned}
   a \sim \pi_{\theta}(a|s) = \mathbb{P}(A_{t}|S_{t} = s;\theta )
  \end{aligned}
  \label{action}
\end{equation}

In DRL, the control policy is approximated by a neural network parameterized by $\theta$ \cite{mnih2013playing}. When the control agent takes the action $a$, a state transition $S_{t+1} = s^{\prime}$ occurs based the system dynamics $f_{\theta}$ (Equation~\ref{environment}), and the control agent receives a reward $R_{t+1} = r$. 

\begin{equation}
  \begin{aligned}
    s'\sim f_{\theta}(s,a) = \mathbb{P}(S_{t+1}|S_{t} = s, A_{t} = a)
  \end{aligned}
  \label{environment}
\end{equation}
\begin{equation}
  \begin{aligned}
   \theta^{*} = \underset{\theta}{\mathrm{argmax}}\, \mathbb{E}_{\pi_{\theta}}[r]
  \end{aligned}
  \label{parameter}
\end{equation}

Due to the Markov property, both reward and state transition depend only on the previous state. DRL then finds a policy $\pi_{\theta}$ that maximizes the expected reward (Equation~\ref{parameter}).

Why do we use DRL for irrigation control?
\begin{itemize}

    \item DRL learns an optimal irrigation control policy directly from data, without using any pre-programmed control rules or explicit assumptions about the soil-water environment. 

    \item DRL allows us to use domain knowledge to train an irrigation control agent (a neural network) without labeled data.
    
    \item The generalization ability of the neural network enables the control agent to better handle the dynamically-varying weather and ET data.
    
\end{itemize}

\begin{figure}[t]
\includegraphics[height=1.2in, width=3.1in]{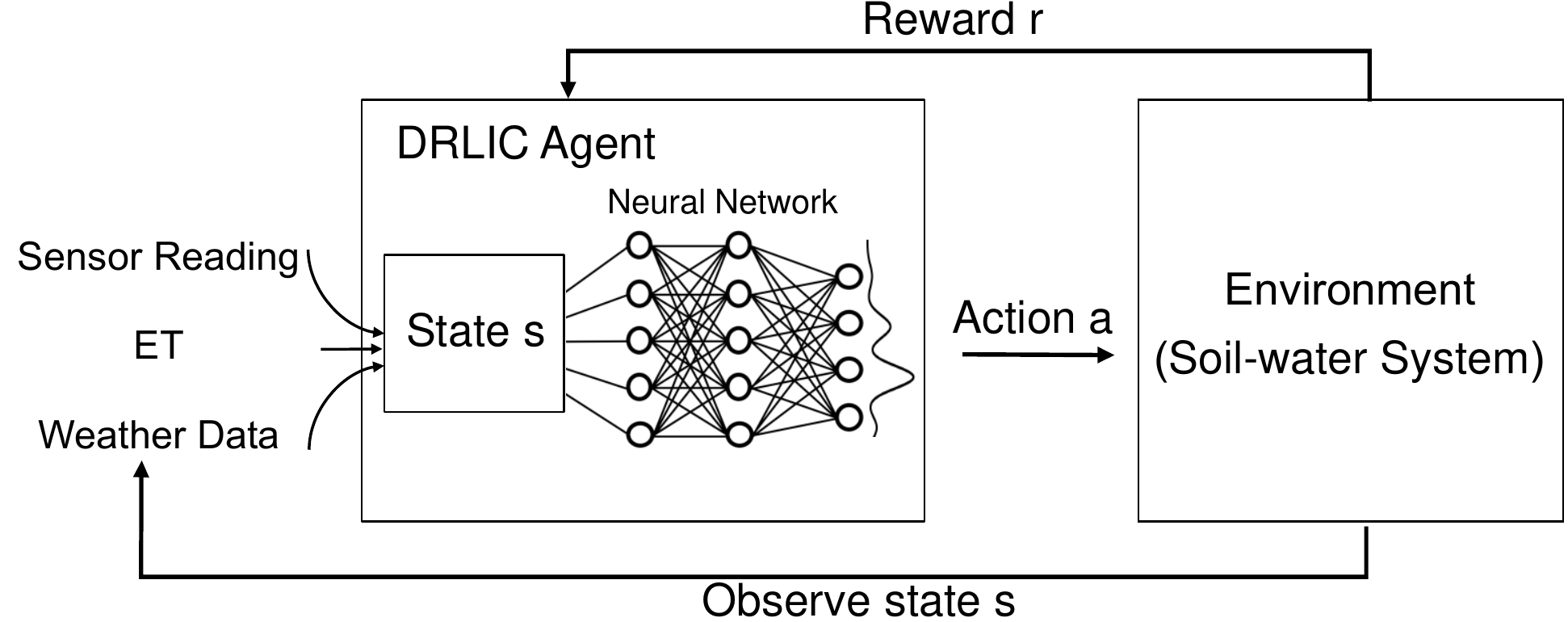}\Description{framework}
  \caption{Deep Reinforcement Learning in \aliasAPP.}
  \label{motivation_rl}
\end{figure}

\subsection{Deep Reinforcement Learning in \aliasAPP}

Figure \ref{motivation_rl} summarizes the DRL architecture of \aliasAPP. 
The irrigation control policy (\aliasAPP Agent) is derived from training a neural network. 
The agent takes a set of information as input, including current soil water content, today's weather data (e.g., ET and precipitation), and the predicted weather data for tomorrow. 
Based on the input, the agent outputs the best action, i.e., the amount of water to irrigate. Until the next day at 11 PM, the resulting soil water content is observed and passed back to the agent to calculate a reward. 
The agent uses the reward to update the parameters of the neural network for better irrigation control performance. 
Next, we introduce the design of each DRLIC component.

\subsubsection{State in \aliasAPP} The state in our irrigation MDP model contains the information of three parts. (a) Sensed state, which is the soil water content measured by \aliasAPP nodes. (b) Weather-related state, which includes the current and predicted state variables from weather station. (c) Time-related state, which is about date information. 

\textit{Sensed State.} The soil water content of each irrigation region can be determined through the use of Equation \ref{water_content}, which takes into account the sensor readings $\varphi$ obtained from the \aliasAPP node.


\textit{Weather-related State.} It is a vector containing the weather information of current day and next day: ET (inch), Precipitation (inch), maximum, average, minimum Temperature ($^{\circ}$F), maximum, average, minimum Humidity (\%), average Solar Radiation (Ly/day), average Wind Speed (mph), Predicted ET by Equation \ref{et_model_equation} (inch), and forecasted Precipitation (inch) from local weather station.

\textit{Time-related State.} Date including the month. There are several states that can change over time, including the water requirements of plants and the weather-related state introduced above. Plant water requirements vary by growth stage, and weather conditions change with the seasons. 
\subsubsection{Action in \aliasAPP}
\label{action_section}
Based on the current state outlined above, our irrigation scheduling is to find the best amount of water to irrigate (inch), which can maintain plant health (or maximize production) with minimum water consumption. 
The action is a vector that contains the water amount to irrigate for each irrigation region in an orchard.
When the agent outputs an action, we will convert the amount of irrigation water to the open time duration (td) $td_{i}$ for i$th$ micro-sprinkler. It is calculated as $td_{i} = a_{i}/I$, where $I$ is the irrigation rate. We set $I$ to 0.018 inch/min according to the specifications of the micro-sprinklers used in our testbed.

\subsubsection{Reward in DRLIC}
\label{reward_section}
We define the reward function to express our objective of achieving good plant health with minimum water consumption. 
Both plant health and water consumption should be incorporated in the reward function. 
As we know from Section \ref{background}, to achieve the maximum production of almond trees, we need to maintain the soil water content between the MAD level and FC level.
We use the soil water content deviation from these two levels as a proxy for plant health.

To minimize water consumption while not affecting the plant health, we consider three situations in the design of the reward, as shown in Equation \ref{reward}. First, when the soil water content ($V_{i}$) for i$th$ irrigation region is higher than the FC ($V_{fc}$) level, the irrigated water is more than the plants' need. In this case, the plants' health is affected by over-irrigated water, and water consumption is too high. Second, when $V_{i}$ is between $V_{fc}$ and $V_{mad}$, the plants are in good health. In this case, we strive to maintain the $V_{i}$ close to $V_{mad}$ to save water, so we give a reward inversely proportional to the water consumption. 
Third, when $V_{i}$ is lower than $V_{mad}$, the plants are under water stress. 
The plants' health is significantly impacted, proportional to the distance between  $V_{i}$  and $V_{mad}$. 

By considering the above three situations, our reward function is defined as follows:

\begin{equation}
R = -\sum_{i=1}^{N}R_{i}
\end{equation}
\begin{equation}
R_{i} =
\left\{
             \begin{array}{lr}
                  \lambda_{1}*(V_{i} - V_{fc} )    +   \mu_{1}*a_{i} ,  &V_{i}> V_{fc}\\
                 \mu_{2}*a_{i} ,  &V_{fc}> V_{i}> V_{mad}  \\
              \lambda_{3}*( V_{mad}  - V_{i} )    +    \mu_{3}*a_{i},  &V_{i}<V_{mad}\\
             \end{array}
\right.
\label{reward}
\end{equation}
\begin{equation}
\begin{array}{l}

V = \sum_{j=1}^{M}\varphi_{j}*d_{j}

\end{array}
\label{water_content}
\end{equation}
\begin{equation}
\begin{array}{l}

V_{mad} = \alpha * V_{awc} + V_{pwp}

\end{array}
\label{lowerbound}
\end{equation}
\begin{equation}
\begin{array}{l}

V_{fc} = V_{awc} + V_{pwp}

\end{array}
\label{upperrbound}
\end{equation}
\begin{equation}
\begin{array}{l}

V_{pwp} = \varphi_{pwp} * D_{inch}

\end{array}
\label{pwp}
\end{equation}
\begin{equation}
\begin{array}{l}

V_{awc} = \sigma_{awc} * D_{foot}

\end{array}
\label{awc}
\end{equation}


\noindent
where $N$ is the number of irrigation regions in one orchard. $a$ is the amount of water from the RL agent. $\sigma_{awc}$ and $\varphi_{pwp}$ are set by referring to the manual of California Almond Board \cite{almond_manual} based on our specific soil type in our testbed. 
Equations \ref{water_content}, \ref{lowerbound}, \ref{upperrbound}, \ref{pwp} and \ref{awc} have been introduced in Section \ref{background}. 

In our current implementation, the parameters of our reward function are set to the values shown in Table \ref{parameter_setting}, based on the specifications of our testbed.
The parameters in Equation \ref{reward} (i.e., $\lambda_{1}$, $\mu_{1}$, $\mu_{2}$, $\lambda_{3}$ and $\mu_{3}$) are set to the best values that provide the best rewards during training. Their values are set by grid search, which will be introduced in detail in Section \ref{tuning}.
The values of these parameters in Table \ref{parameter_setting} confirm with our design goal of the reward function. First, when $V_{i}$ is larger than $V_{fc}$, we give penalties due to both plants' health and water consumption ($\lambda_{1}=3$, but $\mu_{1}=8$). Second, when $V_{i}$ is lower than $V_{mad}$, we give a higher penalty due to plants' health ($\lambda_{3}=10$, but $\mu_{3}=1$).  

\begin{table}[t]
  \renewcommand\arraystretch{0.8}
  \caption{Parameter Setting in Reward.}
\vspace{-0.1in}
  \centering
  \begin{tabular}{c|c||c|c}
    \hline
     \textbf{Parameter} & \textbf{Value} &\textbf{Parameter} & \textbf{Value}
\\
    \hline
   $\lambda_{1}$   & 3  &$\alpha$& 50 (\%)\\
    \hline
   $ \mu_{1}$ &  8 &$D_{inch}$, $D_{foot}$ & 23.62 inches, 1.97 (feet) \\
    \hline
   $\mu_{2}$   & 3 & $d$ &  11.81 (inches)\\
    \hline
   $\lambda_{3}$   & 10& $\varphi_{pwp}$ & 10 (\%)\\
    \hline
   $ \mu_{3}$ &  1 &$ \sigma_{awc}$&  2.4 (in./ft.)\\

    \hline    
  \end{tabular}
  \label{parameter_setting}
\end{table}

\subsection{DRLIC Training} 
\subsubsection{Policy Gradient Optimization}
In the above DRL framework, a variety of policy gradient algorithms can be used to train the irrigation control agent. 
Policy gradient algorithms achieve the objective in~Equation~\ref{parameter} by computing an estimate of the policy gradient and optimizing the objective through stochastic gradient ascent (Equation~\ref{parameter2}).
\begin{equation}
  \begin{aligned}
    \theta \leftarrow \theta + \alpha \bigtriangledown_{\theta }\mathbb{E}_{\pi_{\theta}}[r]
  \end{aligned}
  \label{parameter2}
\end{equation}


In this work, we use proximal policy optimization (PPO) \cite{schulman2017proximal}, which has been successfully applied in many applications such as navigation \cite{molchanov2019sim} and games \cite{berner2019dota}. 
PPO is known to be stable and robust to hyperparameters and network architectures \cite{schulman2017proximal}. 

PPO minimizes the loss function in Equation~\ref{loss}, which is equivalent to maximizing the Monte Carlo estimate of rewards with regularization. The advantage function $ \hat A_{t}$ given by Equation \ref{advantage} is used to estimate the relative benefit of taking an action from a given state. 
\begin{equation}
  \begin{aligned}
L_{PPO}(\theta) = -\hat{\mathbb{E}}_{t}[min(w_{t}(\theta)\hat A_{t}, clip(w_{t}(\theta),1-\epsilon ,1+\epsilon)\hat A_{t})]
  \end{aligned}
  \label{loss}
\end{equation}
\begin{equation}
  \begin{aligned}
    \hat A_{t} = \sum_{i=0}^{\infty }\gamma^{i}r_{t+i}
  \end{aligned}
  \label{advantage}
\end{equation}
\begin{equation}
  \begin{aligned}
w_{t}(\theta) = \frac{\pi_{\theta}(a_{t}|s_{t})}{\pi_{\theta_{old}}(a_{t}|s_{t})}
  \end{aligned}
  \label{importance_weight}
\end{equation}
In Equation \ref{importance_weight}, $\pi_{\theta}(a_{t}|s_{t})$ is the policy being updated with the loss function and $\pi_{\theta_{old}}(a_{t}|s_{t})$ is the policy that was used to collect data with environment interaction. As the data collection policy differs from the policy being updated, it introduces a distribution shift. The ratio $w_{t}(\theta)$ corrects for this drift using importance sampling. The ratio of two probabilities can blow up to large numbers and destabilize training, so the ratio is clipped with $\epsilon$.


\subsubsection{Data Collection and Preprocessing}
On day $t$, the \aliasAPP agent observes a state $s$ (e.g., moisture level), and then chooses an action (water amount). After applying the action, the soil-water environment's state transits to $s_{t+1}$ next day and the agent receives a reward $r$. After that, a data pair ($s_{t},a_{t},r_{t},s_{t+1}$) can be collected. We conduct data normalization by subtracting the mean of the states/action and dividing by the standard deviation. We use 10-year weather data (2010-2020) to generate the data pairs in our dataset, which will be used to train our \aliasAPP agent.

\subsubsection{Training Process}
Ideally, \aliasAPP's control agent should be trained in an orchard of almond trees. A well-trained DRL agent needs 384 years to converge due to the long control interval of irrigation systems. It is impossible to train \aliasAPP agent in an orchard. A feasible solution is to refer to a high-fidelity simulator. However, there are no such simulators available in the soil-water domain. Then we decide to leverage a data-driven simulator to speed up the training process. We employ the soil water content predictor introduced in Section \ref{safe_section} as our soil-water simulator. The simulator allows \aliasAPP to "experience" the weather of 10 years in several minutes.


The training procedure of \aliasAPP is outlined in Algorithm \ref{ppo_algo}. We train \aliasAPP using 1000 episodes and length of an episode as 30 days. For each episode, we can collect 30 training data pair ($s_{t},a_{t},r_{t},s_{t+1}$) under different weather data and leverage Equation \ref{loss} to optimize the objective in Equation \ref{parameter} through stochastic gradient ascent. The training ends once Algorithm \ref{ppo_algo} converges: at the end of each episode the total reward obtained is compared with the previous total. If the current episode reward does not change by ± 3\%, we consider the policy has converged. If the policy does not converge, the training will continue up to a maximum of 100 training iterations (\# episodes = 100). After the training, we will deploy the trained \aliasAPP agent into the real almond orchard.

When we are given a new environment (e.g., a new orchard),  we first need to collect the real-world irrigation data of new environment by existing controller (e.g., ET-based control) to build a soil water content predictor to describe the water balance in the root zone soil. Then we leverage the soil water content predictor to speed up the training process, after that, we deploy the well-trained \aliasAPP agent for this new orchard.

\begin{algorithm}[t]
    \caption{\aliasAPP Training Algorithm}
    \label{ppo_algo}
    \KwIn{
    State s, Action a, Reward r,
    an initialized policy, $\pi_{\theta}$;
    }
    \KwOut{A trained irrigation control agent \;}

    \For{$i=0,..., \#$ Episodes}
    {
        $State \leftarrow  Soil-water environment$\;
        
        $\theta_{old} \leftarrow \theta\ $ \;
     
        \For{t = 0, ...,$\#$ Steps }
        {
          $\hat a_{t} = \pi_{\theta}(s_{t})  $\;
          $s_{t+1}, r_{t+1} = env.step(\hat a_{t})$\;

        }
        
    Compute $ \hat A_{t}$ \;
    With minibatch of size M\;
    $\theta \leftarrow \theta\ - \alpha \triangledown_{\theta}L_{PPO}(\theta)$ \;
    }
\end{algorithm}

\begin{figure}[t]
\includegraphics[height=2.6in, width=3in]{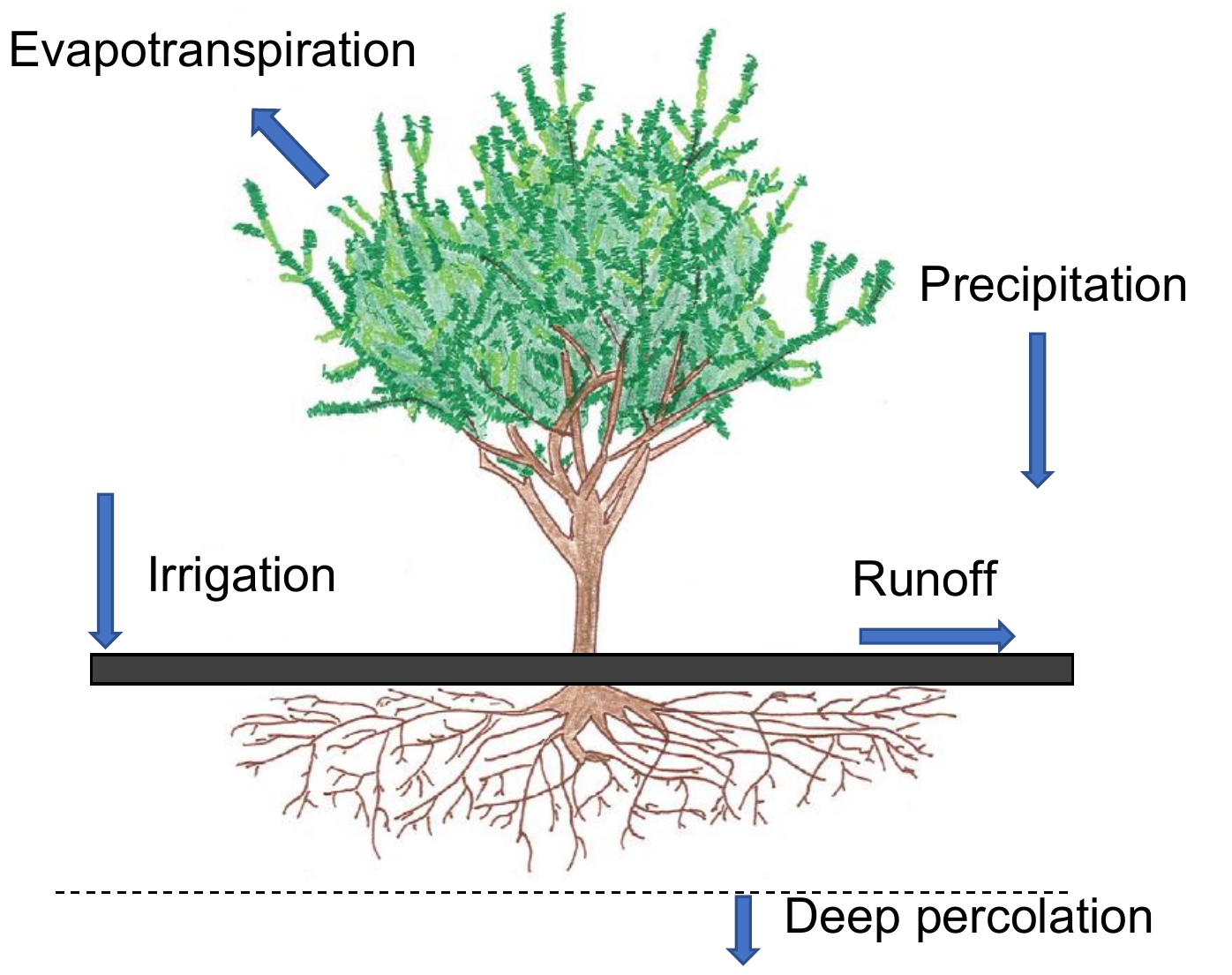}\Description{waterbalance}
  \caption{Water Balance in the Root Zone Soil.}
  \label{waterbalance}
\end{figure}

\subsection{Safe Mechanism for Irrigation} 
\label{safe_section}

\begin{figure}[t]
  \includegraphics[height=1.5in, width=3.3in]{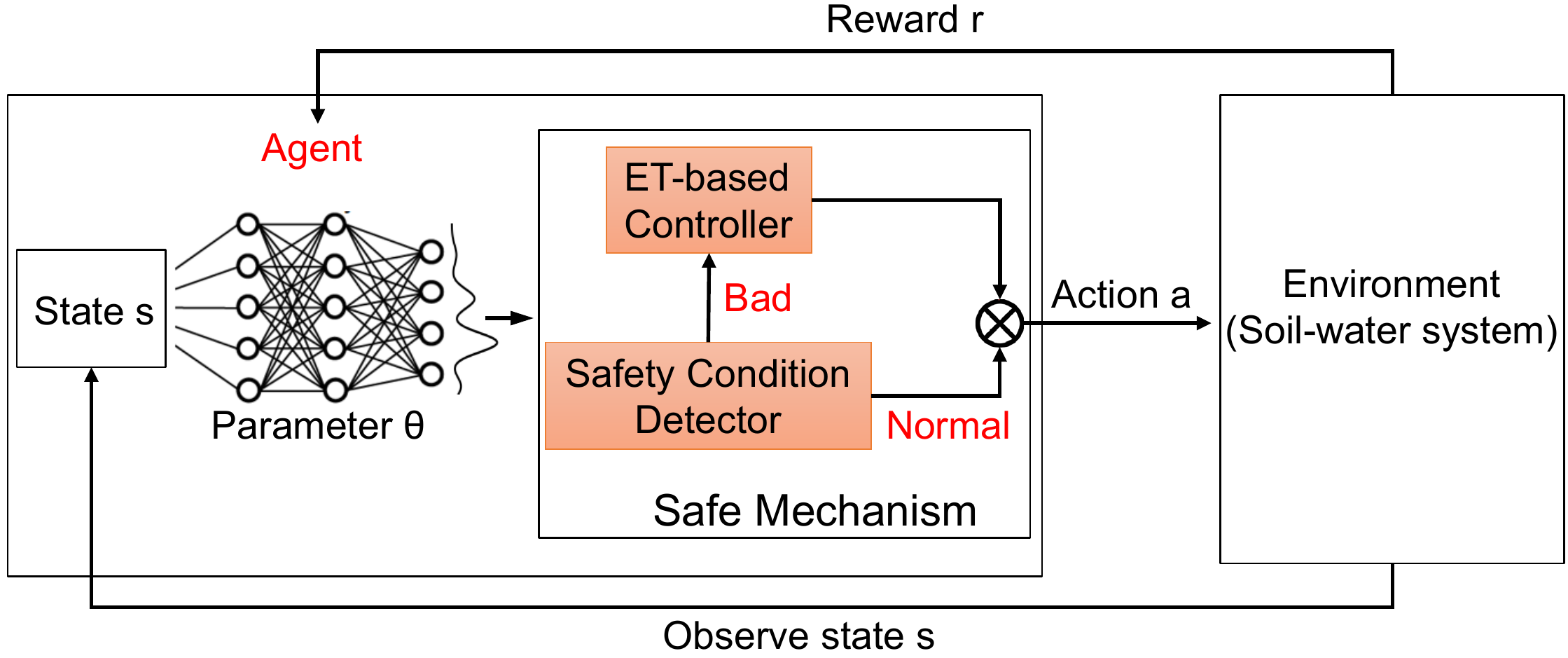}\Description{rl}
  \caption{Reinforcement Learning with Safe Mechanism.}
  \label{on_rl}
\end{figure}

We design a safe mechanism that integrates the RL and ET controller in a coupled close-loop. Figure \ref{on_rl} illustrates the workflow of safe mechanism, with the following key elements. (i) Different from the pure RL framework, we introduce a safety moisture condition detector to evaluate whether the RL algorithm outputs a safe action. (ii) If so, the action goes to the RL agent, who will be in charge of irrigation control. (iii) Otherwise, we will use an ET-based controller to generate an action for that control cycle. (iv) \aliasAPP will the RL agent for the future control cycles. 
We now introduce the soil water content predictor and safety condition detector.

\textbf{Soil Water Content Predictor.} 
To enable early detection of an unsafe action, we design a soil water content predictor to predict the moisture trend after taking an action. Then we design a safe condition detector to detect almond health penalty $p(t)$. The idea is to detect whether the damage metric for an almond tree is higher than a threshold. If so, the detector will command \aliasAPP to switch from RL to ET-based controller.

We design a soil water content predictor to describe the water balance in the root zone soil. As shown in Figure \ref{waterbalance}, the variations of water storage in the soil are caused by both inflows (irrigation and precipitation) and outflows (evapotranspiration). This leads to the following mathematical expression: 

\begin{equation}
\begin{array}{l}
V_{i, t+1} = c_{1} * V_{i, t} + c_{2}*(A_{i, t}+P_{t}) + c_{3}*E_{t}  + b
\end{array}
\label{soil_water_equation}
\end{equation}
\begin{equation}
\begin{array}{l}
E_{t} = \Gamma_{c} * RA * TD^{(1/2)} * (T_{t} + 17.8^{\circ}C)
\end{array}
\label{et_model_equation}
\end{equation}
where $V_{i, t+1}$ denotes the predicted moisture level in the root zone for $i$th irrigation region after taking the action from RL, $E_{t}$ and $P_{t}$ are the plants' ET and the measured rainfall. In time period $t$, and $A_{i, t}$ is the irrigation amount for $i$th irrigation region. $c_{1}$, $c_{2}$, and $c_{3}$ are coefficients. It is assumed in this work that runoff and water percolation are proportional to soil moisture level \cite{ooi2008systems, cui2018infiltration, cui2018prediction} in Equation \ref{soil_water_equation}. All the coefficients can be determined by means of system identification techniques \cite{delgoda2016root}. All variables are
normally expressed in inches.

The weather data can be get from local weather station. For $ET$, we adopt the simple calculation model established in ~\cite{hargreaves1985reference}. As shown in Equation~\ref{et_model_equation}, where $\Gamma_{c}$ is a crop-specific parameter. RA stands for extraterrestrial radiation, which is in the same unit as $E_{t}$. $TD$ denotes the annual average daily temperature difference, which can be derived from local meteorological data, and $T_{t}$ is the average outdoor temperature during the $t$th time period.

\textbf{Safety Condition Detector.}
We employ the difference between predicted moisture level and lower bound as a detector to estimate the almond tree damage. As explained in Section \ref{background}, MAD is the lower bound. Then we use $\sum_{i=1}^{N}(V_{mad}  - V_{i, t+1})$ as a safety condition detector, $V_{i, t+1}$ denotes the predicted moisture level from $i$th irrigation region for $t$ timestep. $V_{mad}$ is the water content lower bound. \aliasAPP will evoke ET-based controller once safety condition detector detects the dangerous irrigation action.

\begin{table}[t]
  \renewcommand\arraystretch{0.8}
  \caption{Coefficients of Predictor for Each Tree.}
  \vspace{-0.1in}
  \centering
  \begin{tabular}{c|c|c|c|c|c|c}
    \hline
      & \textbf{c1}  & \textbf{c2} & \textbf{c3} & \textbf{b} & $\textbf{R}^{2}$ &\textbf{NRMSE}\\
    \hline
    Tree1  & 0.973 & 0.288 & -0.103 & 0.003 & 0.982& 0.062 \\
    \hline
    Tree2   & 0.937 & 0.325 & -0.121 & 0.013 & 0.985& 0.071\\

    \hline
  \end{tabular}
  \label{system_ident}
\end{table}

\textbf{Parameter Learning of our Soil Water Content Predictor.}
We leverage the designed testbed to collect the irrigation amount of almond trees for 2 months. The ET value for each day is collected from a local weather station \cite{et_database} and the moisture level for each tree is collected by the designed \aliasAPP node. Then the linear least square method was applied to estimate the coefficients. $R^{2}$ is used to explain the strength of the relationship between the moisture level and related factors. Normalized root-mean-square error (NRMSE) is used as a goodness-of-fit measure for predictors. The results are shown in Table \ref{system_ident}, we can see that the $R^{2}$ is close to 1 indicating that the irrigation, ET and precipitation have a strong relationship with soil water content for the tree. The NRMSE is less than 0.1 which means that the predictor can achieve accurate prediction for soil water content.

\section{Testbed and Hardware}\label{testbed_hardware}

\begin{figure}[t]
  \includegraphics[height=3.0in, width=2.5in]{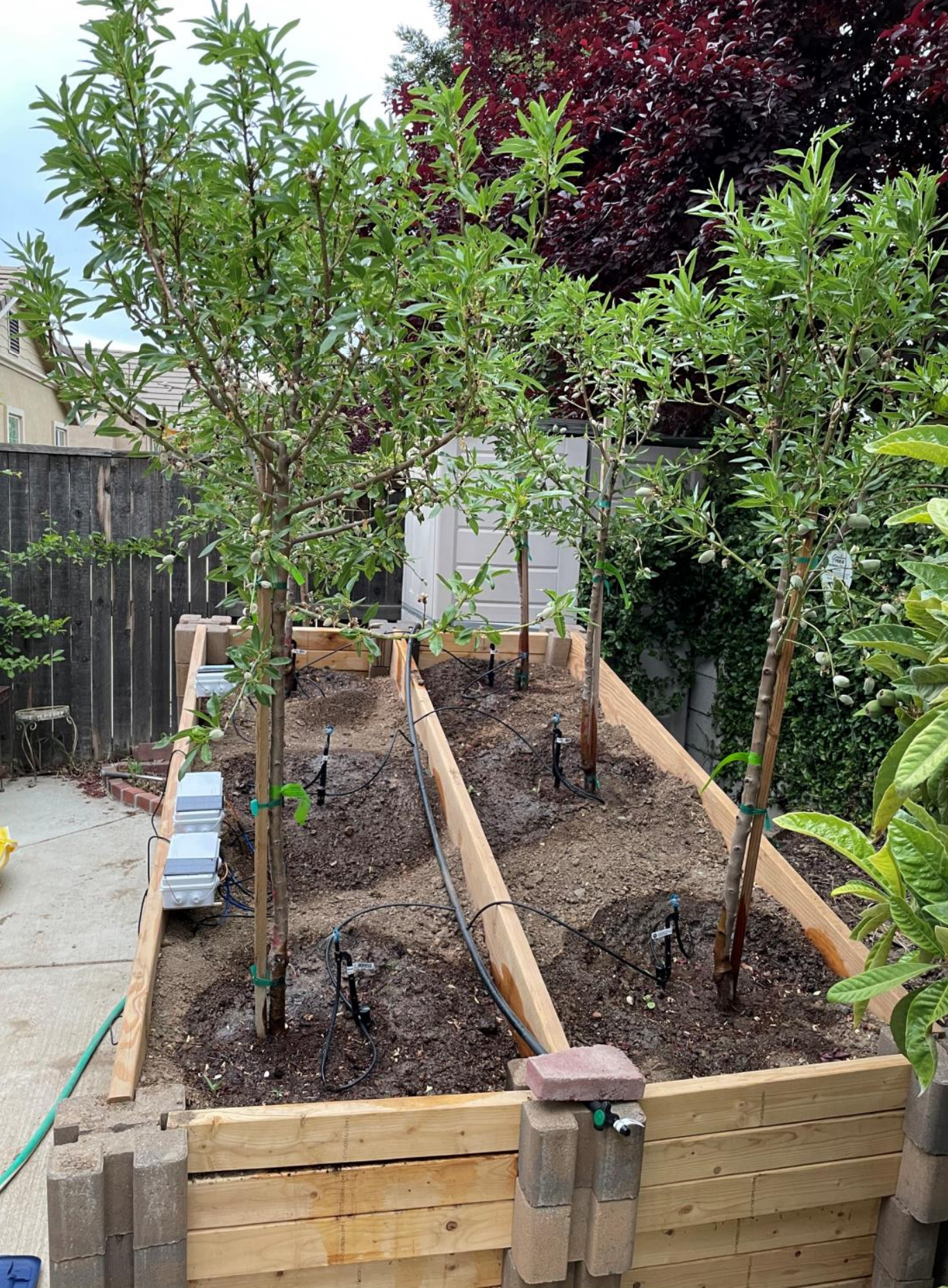}\Description{rl}
  \caption{Testbed and Microsprinkler Irrigation System.}
  \label{test_bed}
\end{figure}

\subsection{Testbed and Microsprinkler Description}
\label{testbed_description}
Figure \ref{test_bed} shows our micro-sprinkler irrigation testbed. The micro-sprinkler irrigation system is installed and designed to be identical in hardware, micro-sprinkler coverage, etc. This irrigation system measures 290 cm x 160 cm, with micro-sprinklers arranged in a 3x2 grid, each 97cm from the next. The micro-sprinklers chosen were 1/4 ', 360 $^\circ$ pattern by Rainbird, which are currently considered
state-of-the-art in micro-sprinkler technology. Six all-in-one young almond trees were planted into the testbed (three for each). The average height is 2 meters. The soil with 2.7 m $^3$ volume is collected from a local orchard that is a typical loam soil and the plant-available water-holding capacity
is 2.4 inches of water per foot. 

\subsection{\aliasAPP Node Development.}


The designed DRLIC node in Figure \ref{framework} consists of four main parts: sensors, actuator, power supply and transmission module.

\textbf{Sensors:} It consists of several moisture sensors for different depths. The moisture sensors vary in their sensitivity and their volume of soil measured. Each moisture sensor for 12-inch depth provides accurate quantitative soil moisture assessment following the Almond Board Irrigation Improvement Continuum \cite{almond_manual}. We assign 2 moisture sensors for each \aliasAPP node since the depth of root zone of the almonds in our testbed is 24 inches.

\begin{figure}[t]
  \includegraphics[height=1.8in, width=3.1in]{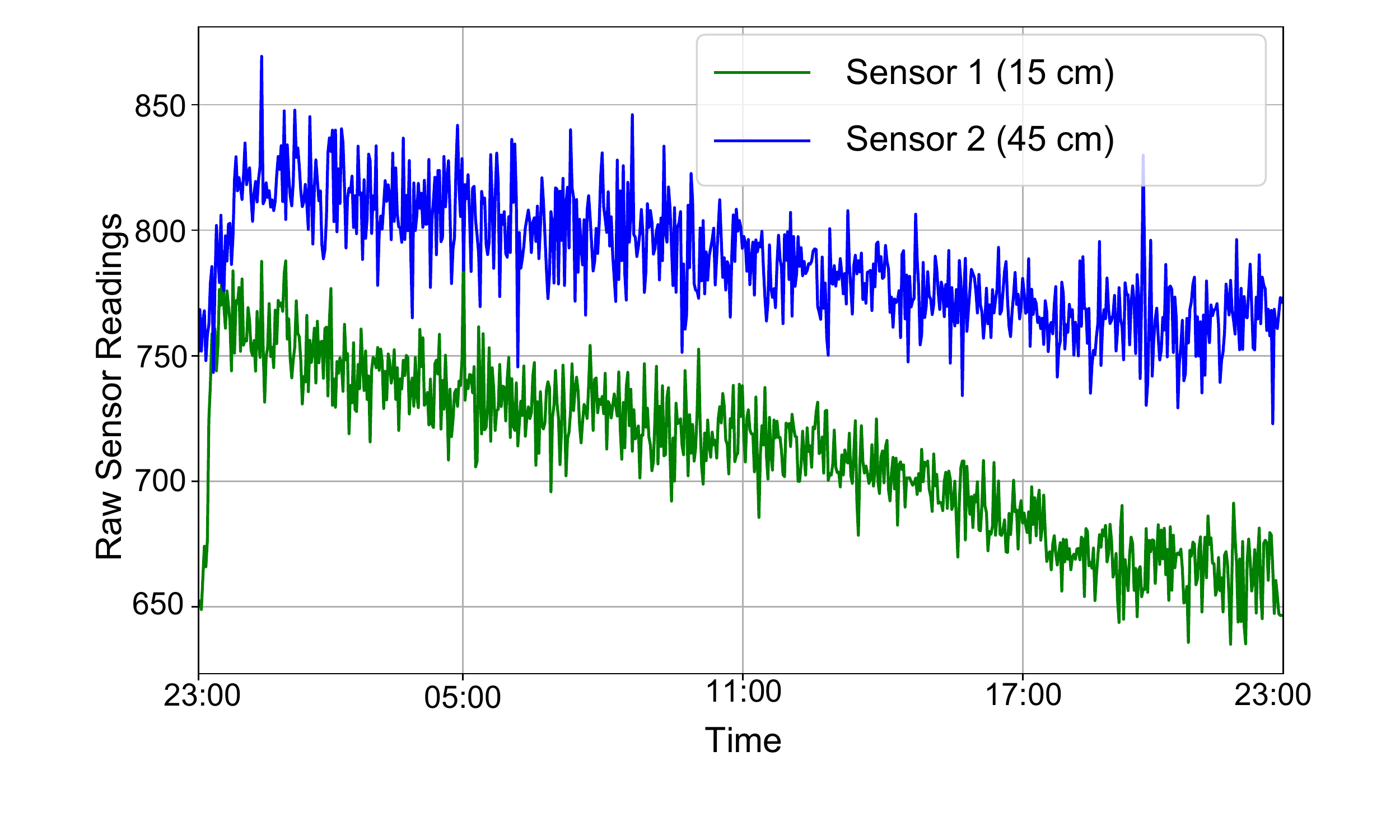}\Description{rl}
  \caption{Daily Soil Moisture Readings.}
  \label{rawreading}
\end{figure}

A key feature of the \aliasAPP node is the ability to measure the volumetric water content in the surrounding soil. We opted to purchase research-quality Decagon EC-5 sensors \footnote{Decagon devices. http://www.decagon.com/products/soils/}, with a reported accuracy of $\pm 3\%$. Raw sensor readings collected over a period of one day with a high sampling frequency can be seen in Figure~\ref{rawreading}. The sensors report the dielectric constant of the soil, which is an electrical property highly dependent on the volumetric water content (VWC).

\begin{equation}
  \begin{aligned}
\varphi (m^{3}/m^{3}) = 9.92 *10^{-4}* raw\_reading - 0.45
  \end{aligned}
  \label{calibration}
\end{equation}

A linear calibration Function \ref{calibration} above provided by the sensor manufacturer is used to convert the raw readings to VWC. The range of $\varphi$ is between 0\% and 100\%. $\varphi$ of saturated soils is generally 40\% to 60\% depending on the soil type.

\begin{figure}[t]
\centering
\subfigure[Positive Current Pulse.]{
\includegraphics[width=3.5cm,height= 4cm]{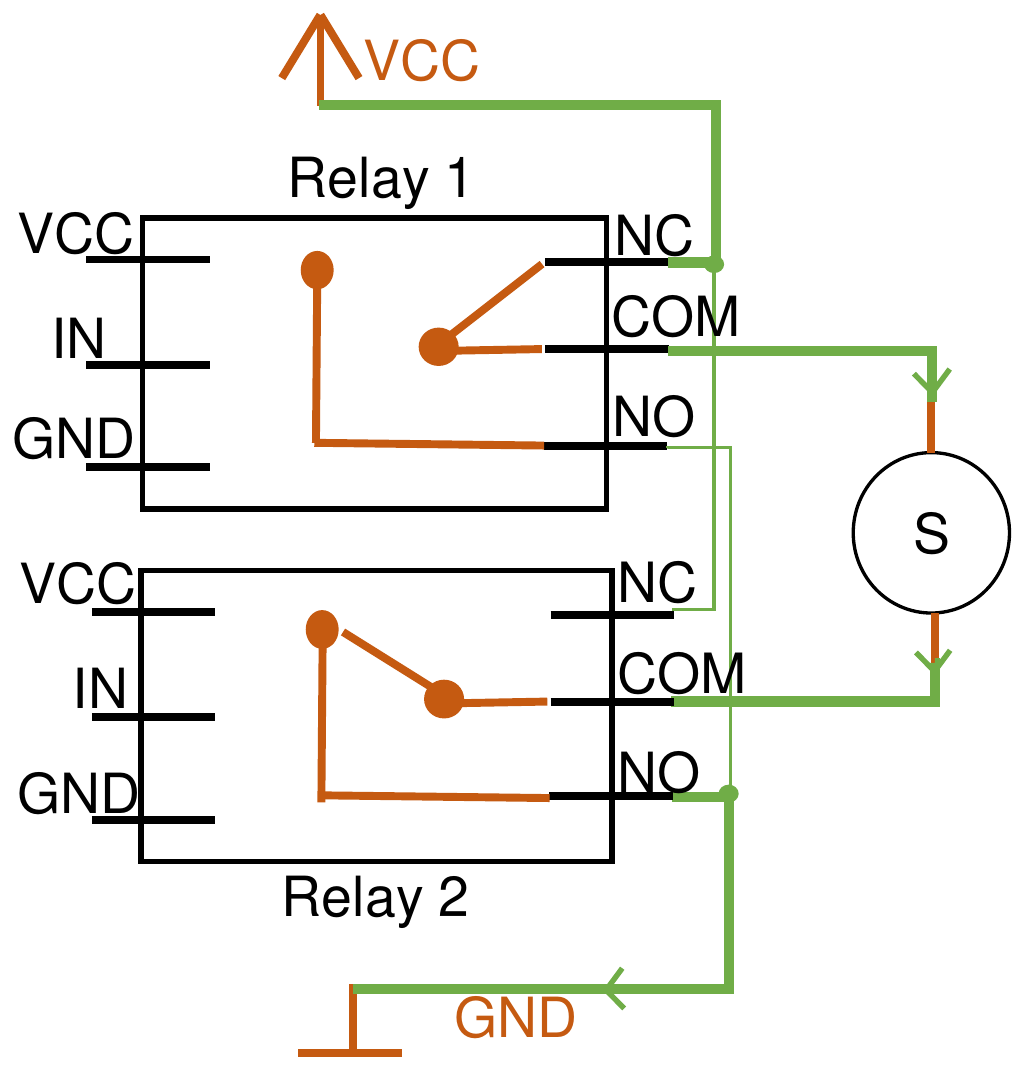}
}
\subfigure[Negative Current Pulse.]{
\includegraphics[width=3.5cm,height= 4cm]{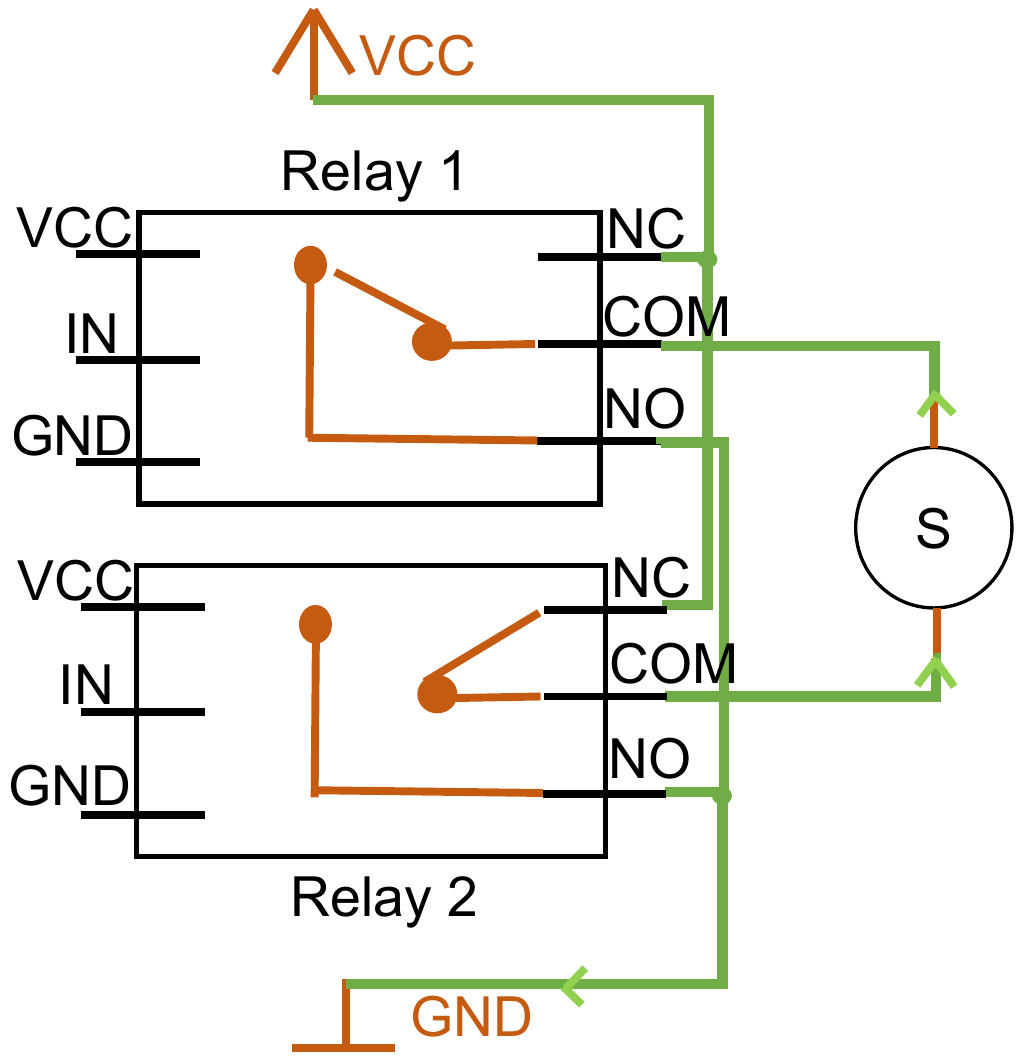}
}
\vspace{-0.1in}
\caption{On and off Circuit Diagram for Latching Solenoid.}
\label{solenoid}
\end{figure}

\textbf{Actuator:} It consists of a latching solenoid with two relays. A standard solenoid requires constant power to allow water to flow, making it a poor choice for a battery-powered system. The nine-volt performance all-purpose alkaline batteries from Amazon can only continue to power the standard 12V DC solenoid for 8 hours. To extend \aliasAPP node lifetime, we chose to use a latching solenoid for micro-sprinkler actuation, requiring only a 25ms pulse of positive (to open) or negative (to close) voltage. The h-bridge is usually used to produce a bi-directional current to control the latching solenoid~\cite{winkler2016magic}. However, it needs a special design to meet different voltages requirements for the ESP32 and latching solenoid. 

In order to control the latching solenoid, we design a circuit diagram using two relays to operate with a very little connection overhead. A relay is an electrically operated switch.   Figure \ref{solenoid} shows the turn-on and off circuit diagram for latching solenoid. When both the relays are off, there is no current going through the solenoid (S). Initially, both the relays are in a normally closed (NC) position. To turn the solenoid on, Relay 1 is switched from NC to normally open(NO) for 25ms, providing the positive current pulse through the solenoid. The current path shown in Figure \ref{solenoid}(a) is: VCC -> NC$_{1}$ -> COM$_{1}$ -> S -> COM$_{2}$ -> NO$_{2}$ -> GND. To turn the solenoid off, Relay 2 is switched from NC to NO for 25ms, de-latching the solenoid to the closed position. The current path shown in Figure \ref{solenoid}(b) is: VCC -> NC$_{2}$ -> COM$_{2}$ -> S -> COM$_{1}$ -> NO$_{1}$ -> GND. To prevent over-irrigation in the event of a power failure, we have the power supply module to continuously provide the power.

\textbf{Power Supply:} Power supply consisted of a 5v, 1.2W solar panel for energy-harvesting and a 18650 Lithium Li-ion battery with a capacity of 3.7V 3000 MAH for energy storage. The TP4056 lithium battery charger module comes with circuit protection and prevents battery over-voltage and reverse polarity connection. All sensors (1 ESP32, 2 moisture sensors, 2 relays and 1 latching solenoid) are powered with this power supply module. It can provide continuous power to prevent over-irrigation in the event of a power failure for the actuator module. 

\textbf{Transmission Module:} Transmission includes uplink and downlink. In the uplink path, the moisture sensor readings from the field are sampled by the ESP32, a low-cost, low-power system on a chip (SoC) series with Wi-Fi capability. The readings are then sent from ESP32 to the base station as input for the optimal control. In downlink path, the control command calculated by the DRL agent will be routed to all ESP32 to turn on or off the solenoids.

\begin{table}[t]
    \renewcommand\arraystretch{1.2}
  \small
  \caption{Hyperparameters}
  \centering
  \begin{tabular}{c|c||c|c}
\hline
\textbf{RL Parameters} & \textbf{Value} & \textbf{General Parameters} & \textbf{Value}\\
\hline
Learning\_rate & 0.01 & Iteration & 1000\\
\hline
Discount\_factor & 0.99 & Number\_neurons & 256\\
\hline
Number\_layers & 2 & Clip\_parameter & 0.3\\
\hline
Minibatches & 128 & Number\_workers & 2\\

\hline
  \end{tabular}
  \label{hyperparameters}
  \label{table: hyperparameters}
\end{table}

\section{Implementation}\label{sec:implementation}
In this section, we illustrate in detail the implementation of \aliasAPP and tuning hyper-parameters.

\noindent
\textbf{\aliasAPP Implementation Details.} We implement the \aliasAPP system in Python, leveraging widely available open-source frameworks like Pandas, Scikit-learn, and Numpy. The control scheme for \aliasAPP is implemented using RLlib \cite{liang2018rllib}, a scalable reinforcement learning framework that supports TensorFlow, TensorFlow Eager, and PyTorch. RLlib provides a range of customizable options for the training process of the \aliasAPP system, including target environment modeling, neural network modeling, action set building and distribution, and optimal policy learning. For our implementation of \aliasAPP, we collected 10 years of weather data (2010-2020) and used 9 years for training and 1 year for testing. We employed the Adam optimizer with a learning rate of 0.01 for gradient-based optimization and set the discount factor to 0.99. The neural network model features 2 hidden layers with 256 neurons each. The local server used for training and running \aliasAPP is a 64-bit quad-core Intel Core i5-7400 CPU at 3.00 GHz running Ubuntu 18.04.

\noindent
\textbf{Training Details and Tuning Hyper-parameters.}
\label{tuning}
The performance of the DRLIC agent is highly dependent on the values chosen for its hyperparameters. However, there is no straightforward approach that guarantees a specific value will improve the total reward obtained by the system. To optimize the DRLIC agent's hyperparameters and improve its performance, we employ a tuning approach that involves optimizing parameters such as $\lambda, \mu$ associated with rewards and penalties, and learning rates. Specifically, we use a grid search approach that allows us to specify a range of values to be considered for each hyperparameter. The grid search process constructs and evaluates the model using every possible combination of hyperparameters. To evaluate each learned model, we employ cross-validation. This tuning approach allows us to identify the best hyperparameter configuration for the DRLIC agent, ensuring optimal DRLIC system performance.




\section{Evaluation}\label{sec:evaluation}
In this section, we evaluate the performance of \aliasAPP in the field. We evaluate \aliasAPP system for 15 days in the real world. 

\begin{figure*}[htbp]
\centering
\subfigure[ET-based Method]{
\includegraphics[width=5.7cm]{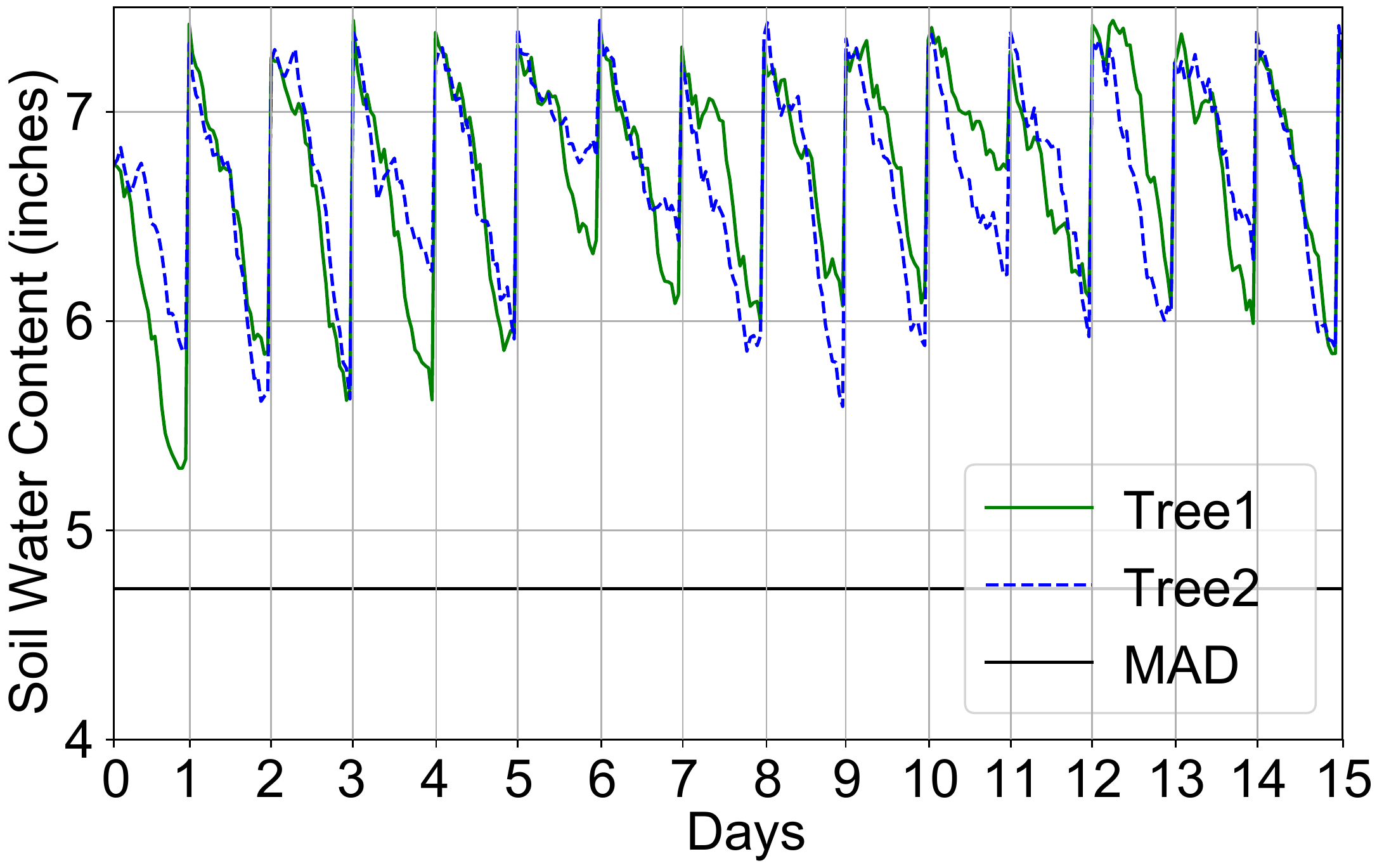}
\label{real_et_health}
}
\subfigure[Sensor-based Method]{
\includegraphics[width=5.7cm]{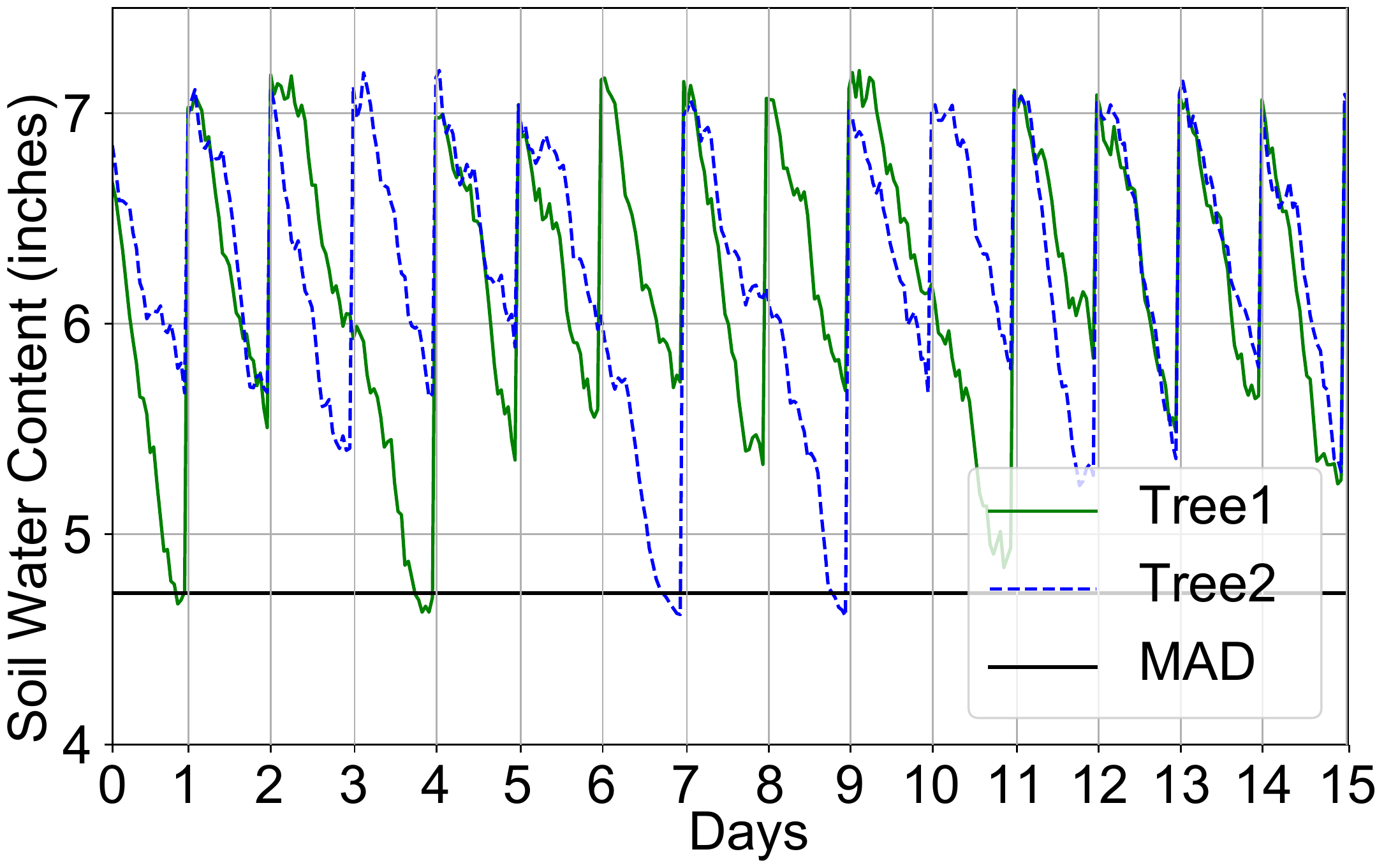}
\label{real_sensor_health}
}
\subfigure[\aliasAPP]{
\includegraphics[width=5.7cm]{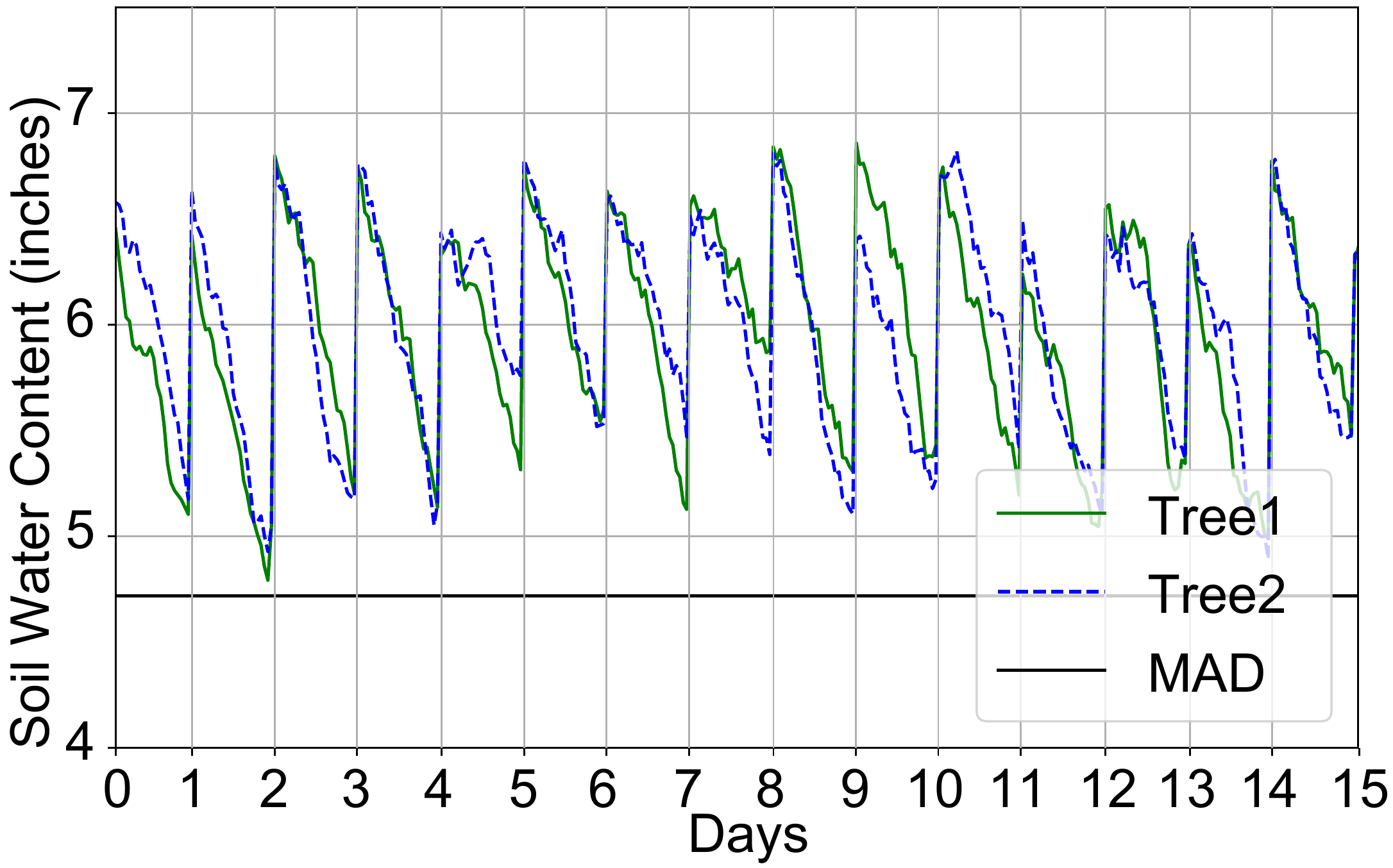}
\label{real_rl_health}
}
\vspace{-0.15in}
\caption{Daily Soil Water Content of Different Irrigation Methods (15 Days).}
\label{plant_health}
\end{figure*}



\subsection{Experiment Setting}

\subsubsection{Baseline Strategy:}
We compare \aliasAPP to two state-of-the-art irrigation control schemes introduced in Section \ref{relate_work}. 

\textbf{ET-Based Irrigation Control \cite{almond_manual}.} 
To implement an ET-based controller, we query a local weather station for the previous day’s ET loss. To compensate for the loss, we use the sprinkler's irrigation rate provided by its dataset to calculate how long the system should be activated for irrigation. 

\textbf{Sensor-based Irrigation Control \cite{grabow2013water}.}
The sensor-based controller has two thresholds, the lower and upper soil water content levels.
The first is set at 4.96 inches, 10\% higher than MAD to avoid the under irrigation occurring prior to the wetting front arriving at the sensor depth. The latter is set to 6.97 inches, 5\% below FC to allow for some rainfall storage. We carefully set these two thresholds based on the soil environment of our testbed.

\subsubsection{Performance Metrics}
\label{performance_metrics}
We evaluate the performance of \aliasAPP and two baseline systems in terms of two performance metrics.

\textbf{Quality of Service.} Although the irrigation system has no control over solar exposure and soil nutrients, it has direct control over the moisture levels in the soil. For this reason, our primary metric for irrigation quality is the system’s ability to maintain soil moisture above this MAD threshold at all times at all of our measured locations. By doing so, we are guaranteeing that the plant has sufficient moisture to be healthy and no production loss. In this paper, we call this the quality of service
of the irrigation system.

\textbf{Water Consumption.} As each sprinkler uses a water supply and we directly control the times at which each micro-sprinkler is active, we can monitor the amount of water consumed by these three systems at all times to determine the efficiency of each system. Thus another metric is the water consumption, which we would like to minimize subject to the quality of service constraints.

\subsubsection{Experiments in our Testbed}
We validate the \aliasAPP system with baselines in real-world deployment in terms of plant health and water consumption for 15 days. In the case study, we have six almond trees in our testbed as shown in Figure \ref{test_bed}. \aliasAPP, sensor-based control and ET-based control are used to irrigate the upper, middle and lower two trees separately since there is no runoff between trees in our testbed. To allow three irrigation systems to operate independently, Every micro-sprinkler is controlled by a \aliasAPP node.  In this way, the only difference among the three systems is the schedules sent to the nodes.


\subsection{Experiment Results}
\begin{figure}[t]
  \includegraphics[height=1.8in, width=3in]{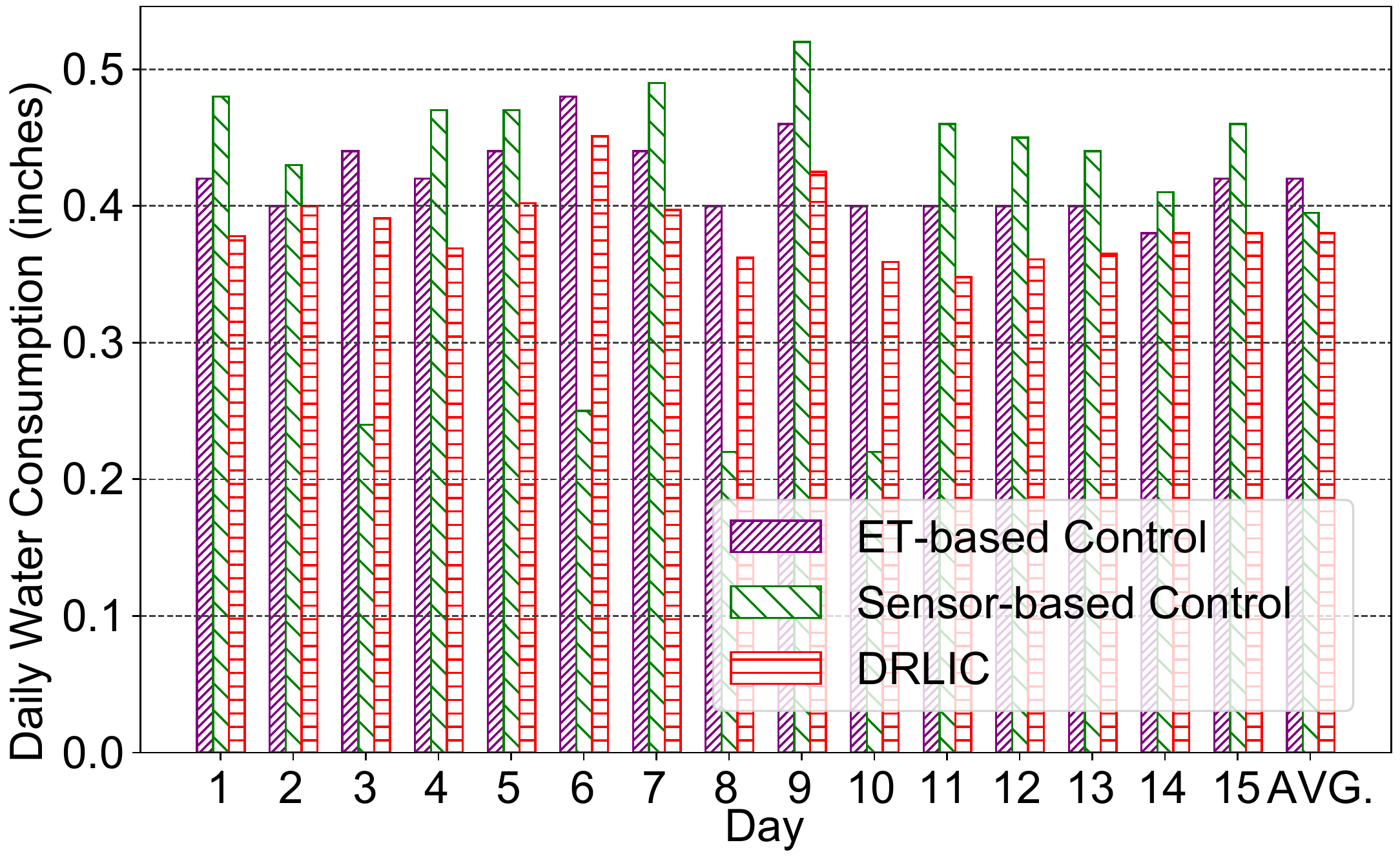}\Description{rl}
    \vspace{-0.15in}
  \caption{Daily Water Consumption.}
  \label{real_water_consumption}
\end{figure}

\subsubsection{Quality of Service}

Irrigation systems are installed to maintain almond health with no production loss.
Figure \ref{real_et_health}, \ref{real_sensor_health}, and \ref{real_rl_health} shows the daily soil water content in the field for ET-based control, Sensor-based control and  \aliasAPP. The black horizontal line shows the MAD level. If soil water content is below this line, tree health will be impacted. We can see that \aliasAPP and ET system can maintain the soil water content above MAD threshold during the 15 days deployment and thus meet the requirement of almond health. However, the trees irrigated by Sensor-based method are in an under-irrigation period of 18 hours for four days (day 1, 4, 7 and 9) since the soil water content of Sensor-based method is lower than the MAD. The reason is that the moisture level of previous day is close but not reached to MAD, so the sensor-based method will not irrigate even though the moisture level is in an under-irrigation trend. \aliasAPP system can irrigate what the trees need based on the learned model about the water changes in the soil and maintain the soil water content close to MAD level.

All three underlying irrigation systems begin with enough water content on the first day. We see that the soil water contents of the two trees in ET control system are much above the FC threshold. 
In our deployment of \aliasAPP against the ET control strategy in Figure \ref{real_et_health}, we see that soil water content for these two trees is different and much higher than the MAD level. This emphasizes the limitations of ET and the core of our work. The irrigated regions don’t receive moisture the same way, and most of the time, the ET-based controller irrigates more water than the plant needs. 



\subsubsection{Water Consumption}
When a decision must be made to switch to a new almond irrigation control system, a primary concern is the efficiency of the proposed system. The system’s ability to return its investment based on increased efficiency will often dictate the acceptance of the technology. In addition, the environmental benefits of reduced freshwater consumption are clear and help promote system adoption. 

In our experimental setup, the water source provided by each micro-sprinkler is pressure-regulated to the industry standard, 30 psi. Each micro-sprinkler head distributing water uses a clearly-defined amount of water per unit time, as described in the almond irrigation manual \cite{almond_manual}. By tracking exactly when each micro-sprinkler is actuated by the system, we can determine very accurately how much water has been consumed. 

Figure \ref{real_water_consumption} shows the daily irrigation amount of two trees actuated by ET-based control, sensor-based control and \aliasAPP in a 15 days' deployment experiment. From this figure, we can see that \aliasAPP can save an average 9.52\% and 3.79\% of the water compared with ET-based and Sensor-based control during 15 days deployment experiment. ET-based control is a centralized control method to irrigate all almond trees without considering their specific need. Sensor-based control is water-efficient by monitoring the moisture and irrigating when the moisture level is lower than the MAD level. However, the thresholds are site-specific and not optimal. \aliasAPP can learn optimal irrigation control by interacting with the local weather and soil water dynamic environment.

\begin{figure*}[htbp]
\centering
\subfigure[ET-based Method]{
\includegraphics[width=5.7cm]{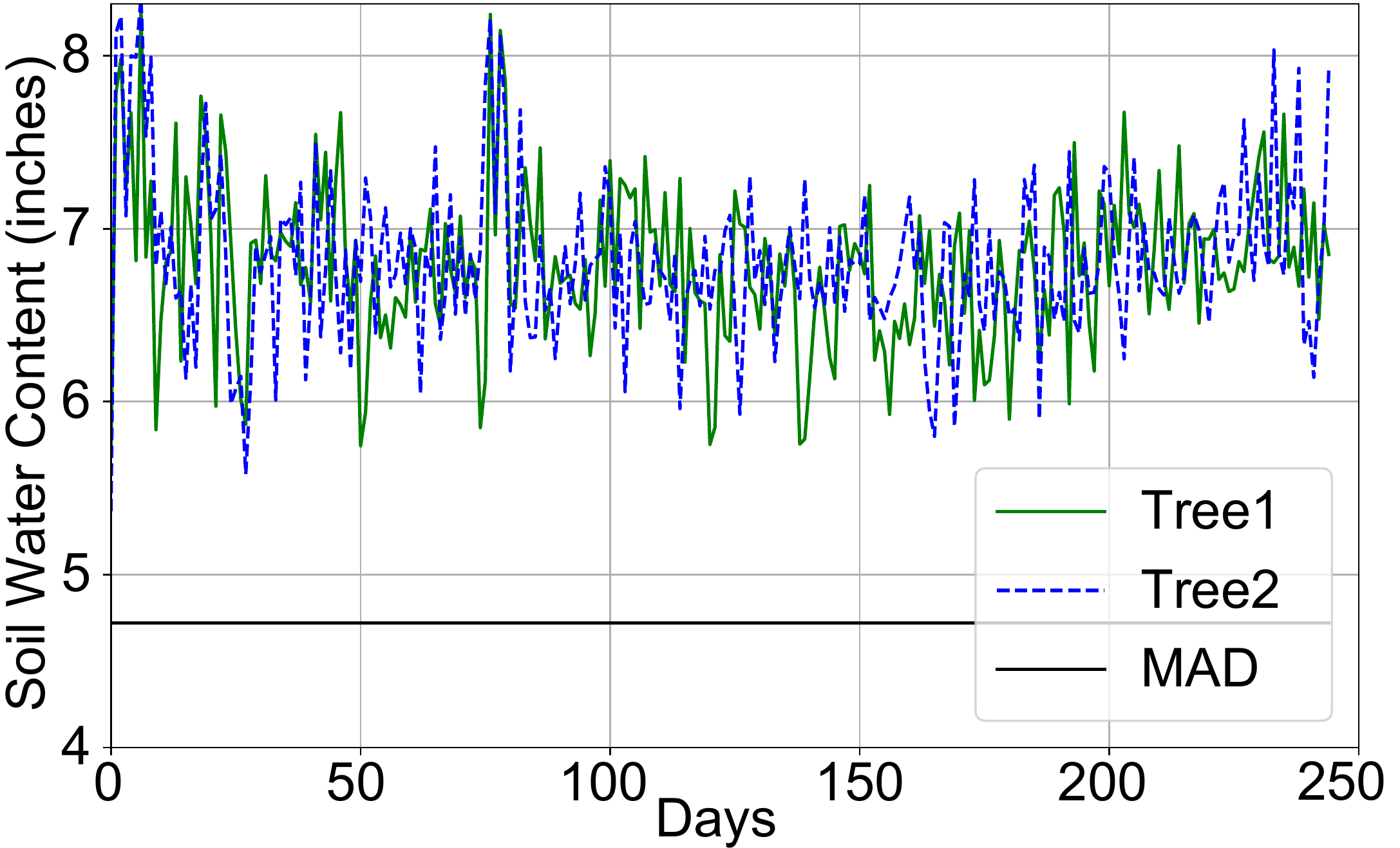}
\label{simulation_et_health}
}
\subfigure[Sensor-based Method]{
\includegraphics[width=5.7cm]{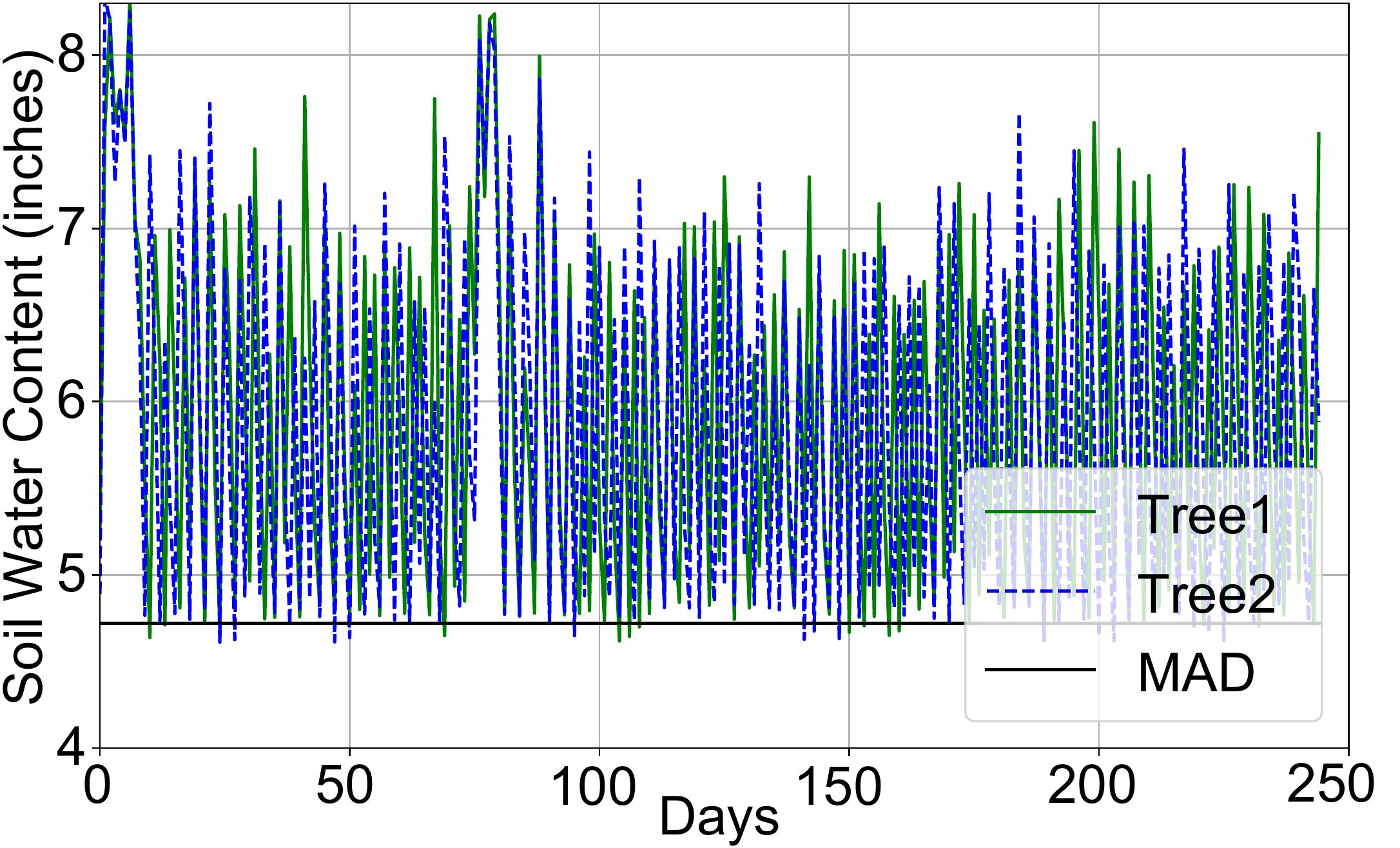}
\label{simulation_sensor_health}
}
\subfigure[\aliasAPP]{
\includegraphics[width=5.7cm]{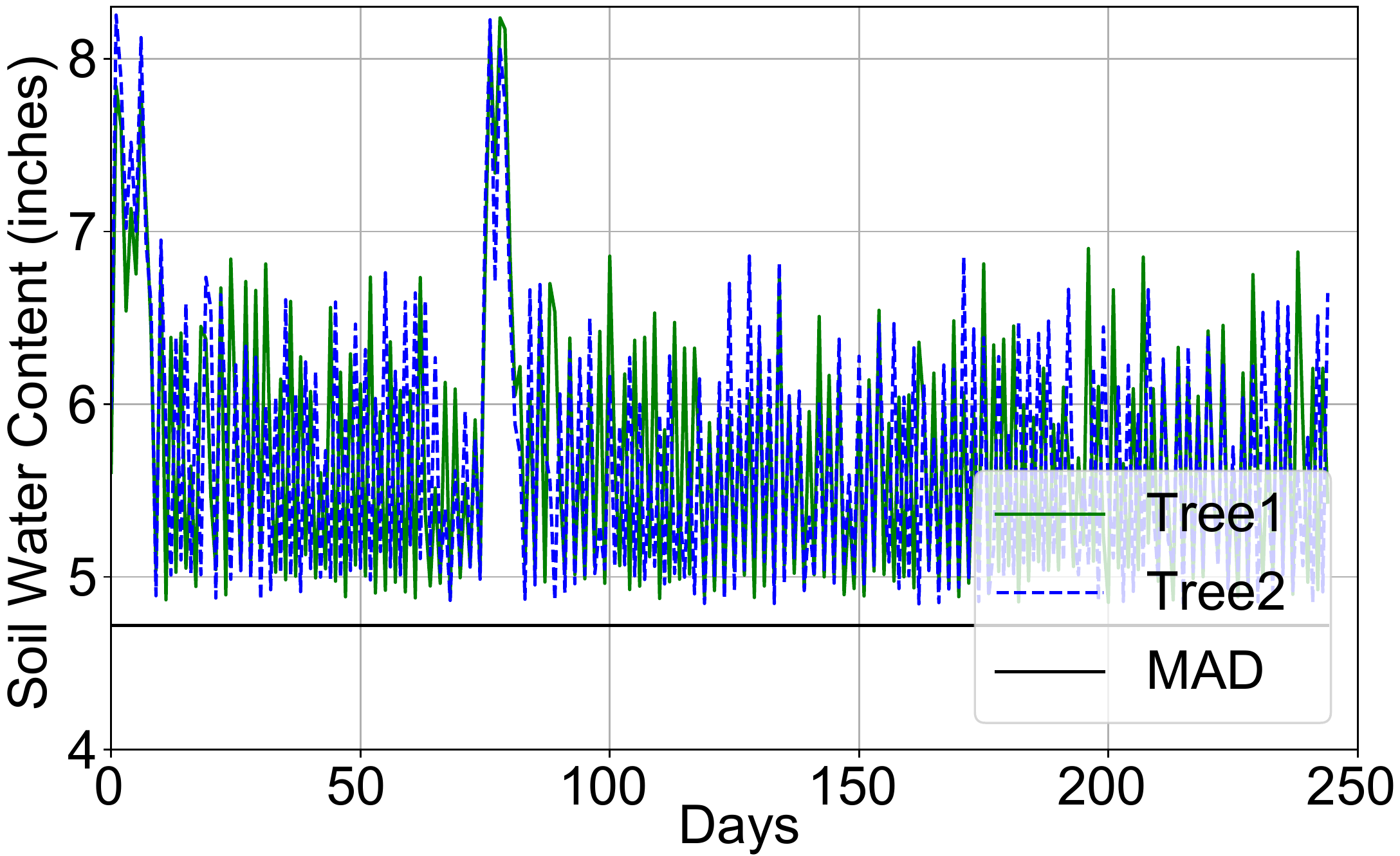}
\label{simulation_rl_health}
}
\vspace{-0.1in}
\caption{Daily Soil Water Content of Different Irrigation Methods.}
\label{simulation_plant_health}
\end{figure*}


 \begin{figure*}[t]
	\begin{minipage}[t]{0.47\linewidth}
		\centering
		 \includegraphics[width=3.0in,height=1.8in,angle=0]{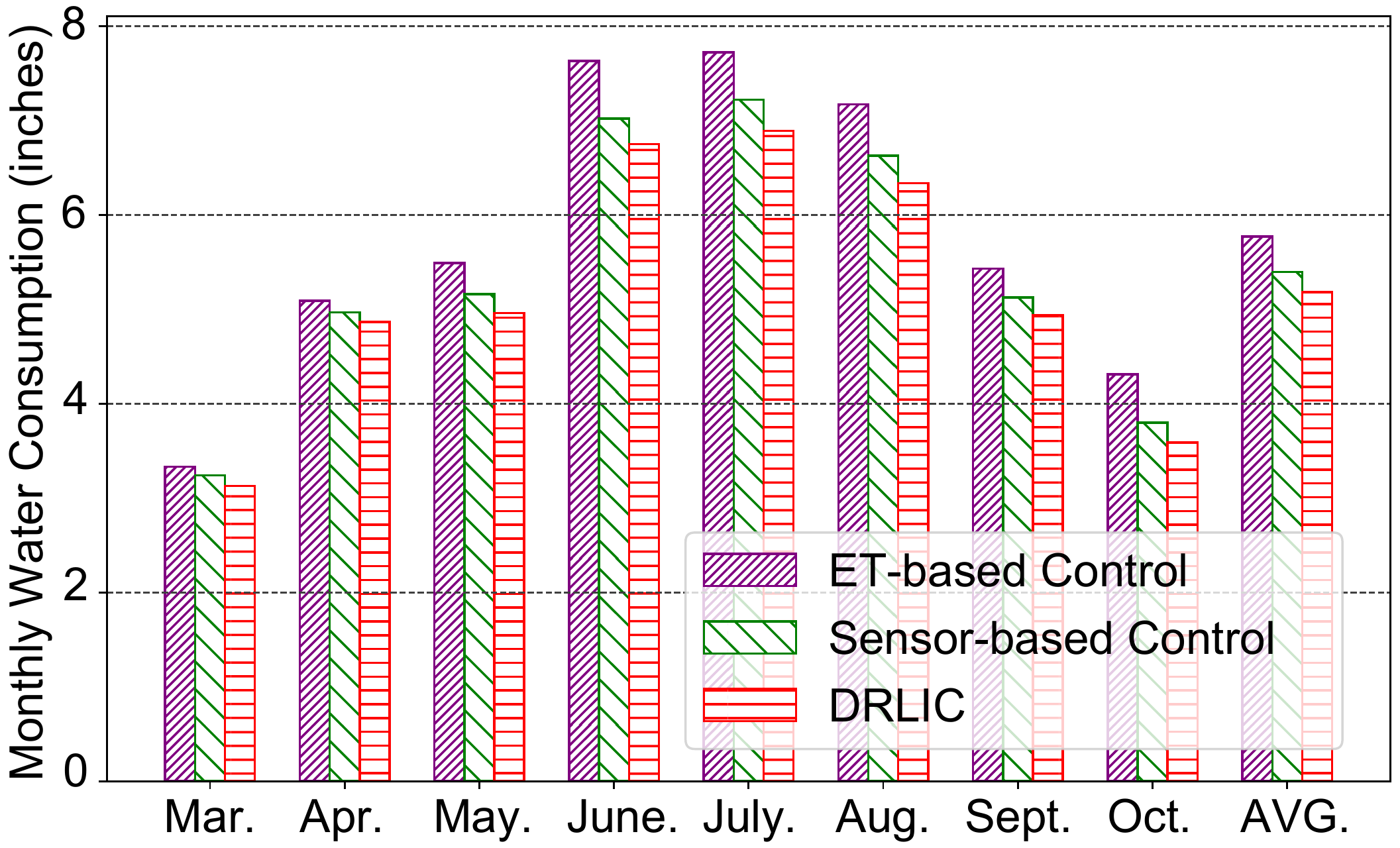}
		\caption{Monthly Water Consumption.}
		\label{month_water}
	\end{minipage}%
		\hspace{1ex}
		\begin{minipage}[t]{0.47\linewidth}
		\centering
		 \includegraphics[width=3.0in,height=1.8in,angle=0]{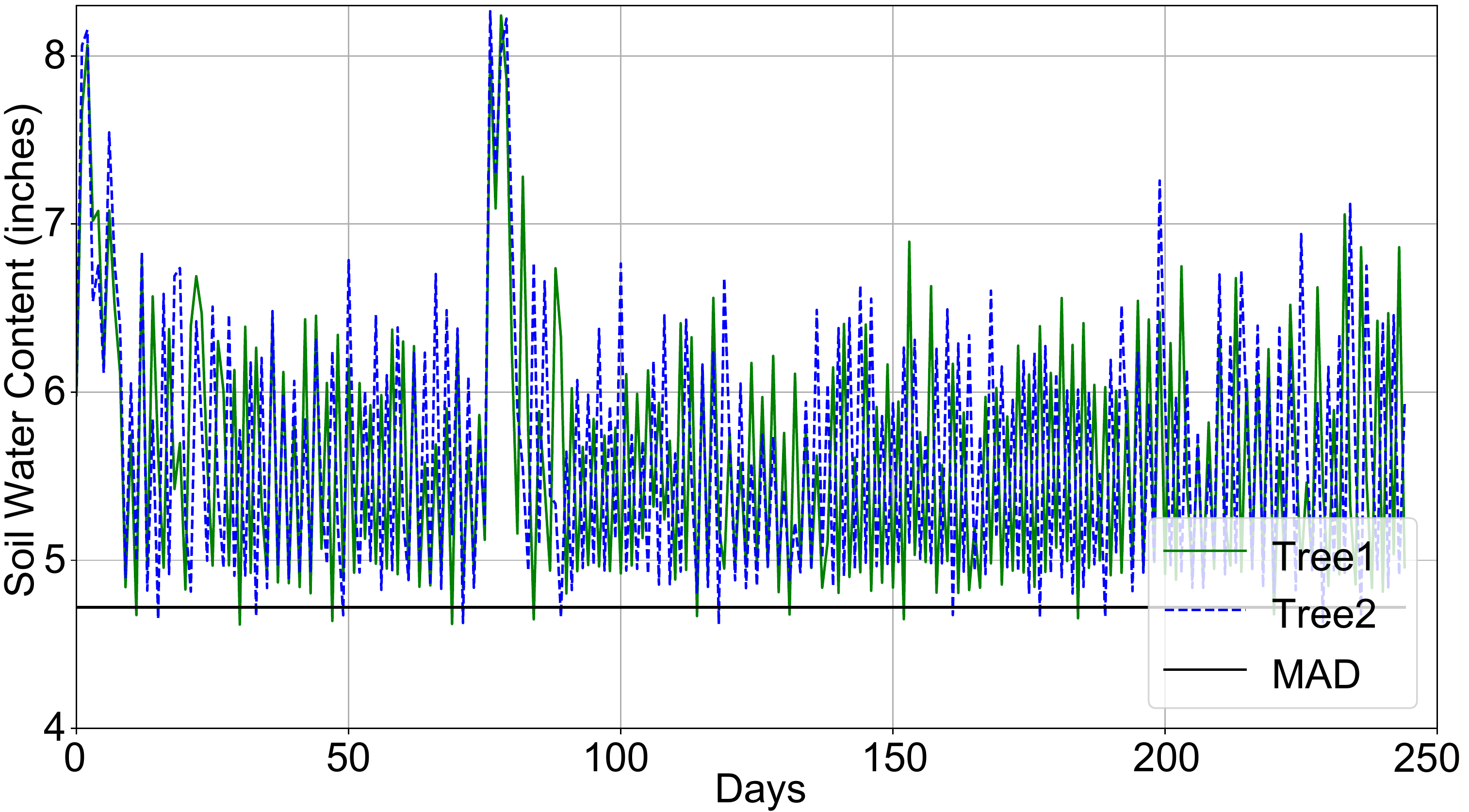}
		\caption{Daily Soil Water Content (w/o Safe Mechanism).}
		\label{rl_no_robust}
	\end{minipage}

\end{figure*}

\subsection{Simulation Results}

In this section, we discuss the simulation results of \aliasAPP and two baselines for a whole growing season.


\subsubsection{Quality of Service}
Figure \ref{simulation_plant_health} shows daily soil water content for ET-based control, Sensor-based control and \aliasAPP in the simulation. The black horizontal line shows the MAD level. We can see that all four control methods can maintain the soil water content above the MAD level and thus meet the requirement of the almond health. Especially on days 2, 5, 24 of March and 16 and 18 of May, we can see that all water content of each tree for all four control systems are the same. The reason is that all of those days have big rainfall and thus make the soil reach the field capacity. In our deployment of \aliasAPP against the ET-based control strategy in Figure \ref{simulation_et_health}, we see that soil water content for these two trees in the control system is different and much higher than the MAD level. 
The sensor-based control maintains two moisture levels, the lower level to start and a higher level to stop irrigation. From Figure \ref{simulation_sensor_health}, we can see that even though we carefully set these two values for each tree,  there are still 43 days the soil water content is slightly under the MAD level which may affect the almond production. Overall, $\aliasAPP_{MAD}$ and \aliasAPP system (Figure \ref{simulation_rl_health}) can maintain the health of all two trees in a uniform irrigation way in the growing season.

\subsubsection{Water Consumption}
Figure \ref{month_water} shows the monthly water consumption by ET-based system, Sensor-based system, and \aliasAPP system from March to October. We can see that \aliasAPP consume less water for each month compared to ET-based and Sensor-based systems. We can also see that the irrigation amount increases and then decreases for all systems. There are two reasons. First, spring rains stop and the weather heats up. Second, the almonds and the trees go through 3 stages in their annual lifecycle, including dormancy during winter, the stunning bloom in March, “growing up” through the spring, and “cracking open” in summer. The year finishes up with harvest spanning from mid-August to October. So in the summer, water consumption is higher than other seasons. Overall, \aliasAPP system can save 10.21\% and  3.93\% of water than ET system and Sensor-based system for the whole growing season. California’s 2019 almond acreage is estimated at 1,530,000 acres, and almond irrigation is estimated to consume roughly 195.26 billion gallons per year \cite{fulton2019water}. With \aliasAPP system, 19.94 and 7.67 billion gallons of water can be saved per year.

\subsection{Effect of our Safe Irrigation Mechanism.}
In the 15 days' deployment, we find that there are two days (Day 2 and 14 in Figure \ref{real_rl_health})
\aliasAPP triggers the ET-control method. This can also be validated from Figure \ref{real_water_consumption}, we can find the water consumption of ET method and  \aliasAPP on days 2 and 14 are the same. We check the weather data to understand the reason and find that the wind speeds of days 2 and 14 are 7.2 and 11.9 mph receptively which is much higher than the average 2.8 mph of the other 13 days.

We now run \aliasAPP with and without safe mechanism for a whole growing season in simulation, labeled as Robust-RL and RL-only, respectively. Figure \ref{simulation_rl_health} and \ref{rl_no_robust} show the daily soil water content of Robust-RL and RL-only for a same growing season 2020, respectively. From the almond's perspective, Robust-RL maintains health with 0 days below the MAD level. The RL-only irrigation method has 21 days below the MAD level. The reason is that the RL models trained from past weather data “misbehave” on the test weather data. while it may be possible to train on changing weather to obtain a robust policy, no offline training can ever cover all possible weather changes. The RL agent with safe mechanism from \aliasAPP,  however, is robust to weather changes because the safety condition detector will detect the dangerous actions from RL agent and the ET system will take control.

 \begin{figure*}[t]
		\begin{minipage}[t]{0.32\linewidth}
		\centering
		 \includegraphics[width=2.2in,height=1.35in,angle=0]{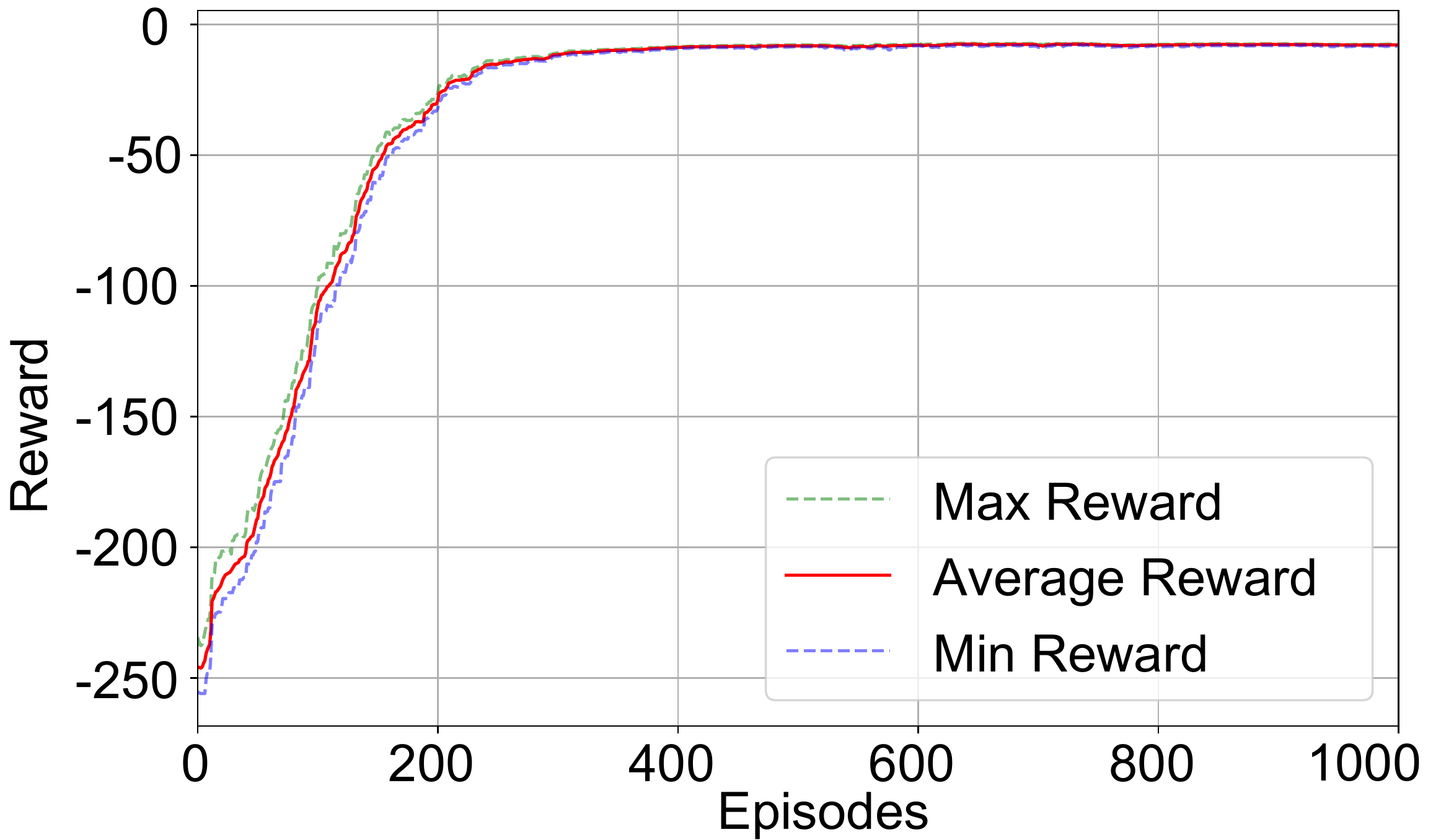}
		\caption{Reinforcement Learning Policy Convergence.}
		\label{reward_converage}
	\end{minipage}	
		\hspace{1ex}
	\begin{minipage}[t]{0.32\linewidth}
		\centering
		 \includegraphics[width=2.2in,height=1.35in,angle=0]{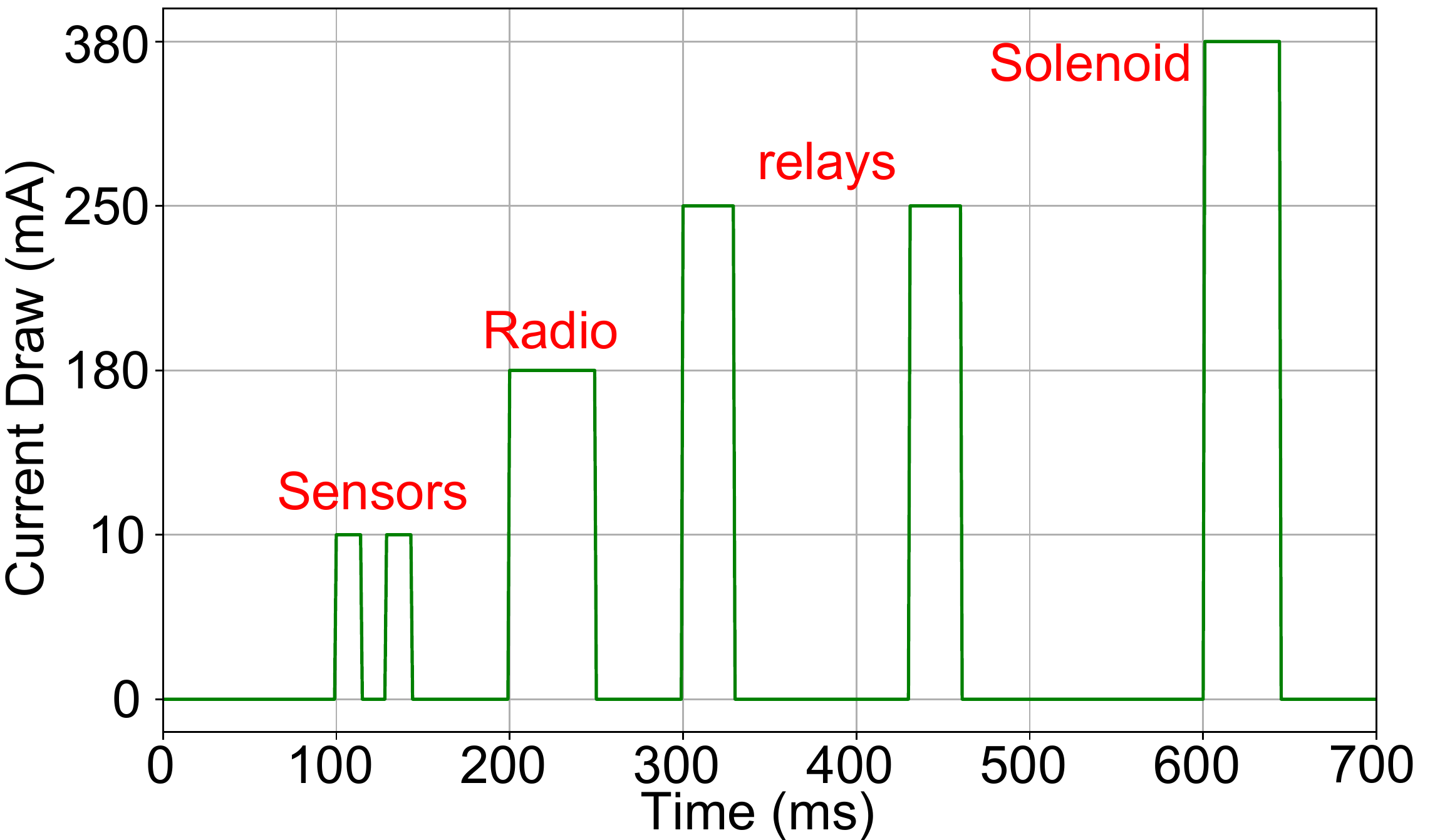}
		\caption{Energy Profile for Different Kinds of Sensors.}
		\label{energy_parts}
	\end{minipage}	
	\hspace{1ex}
	\begin{minipage}[t]{0.32\linewidth}
		\centering
		\includegraphics[width=2.2in,height=1.35in,angle=0]{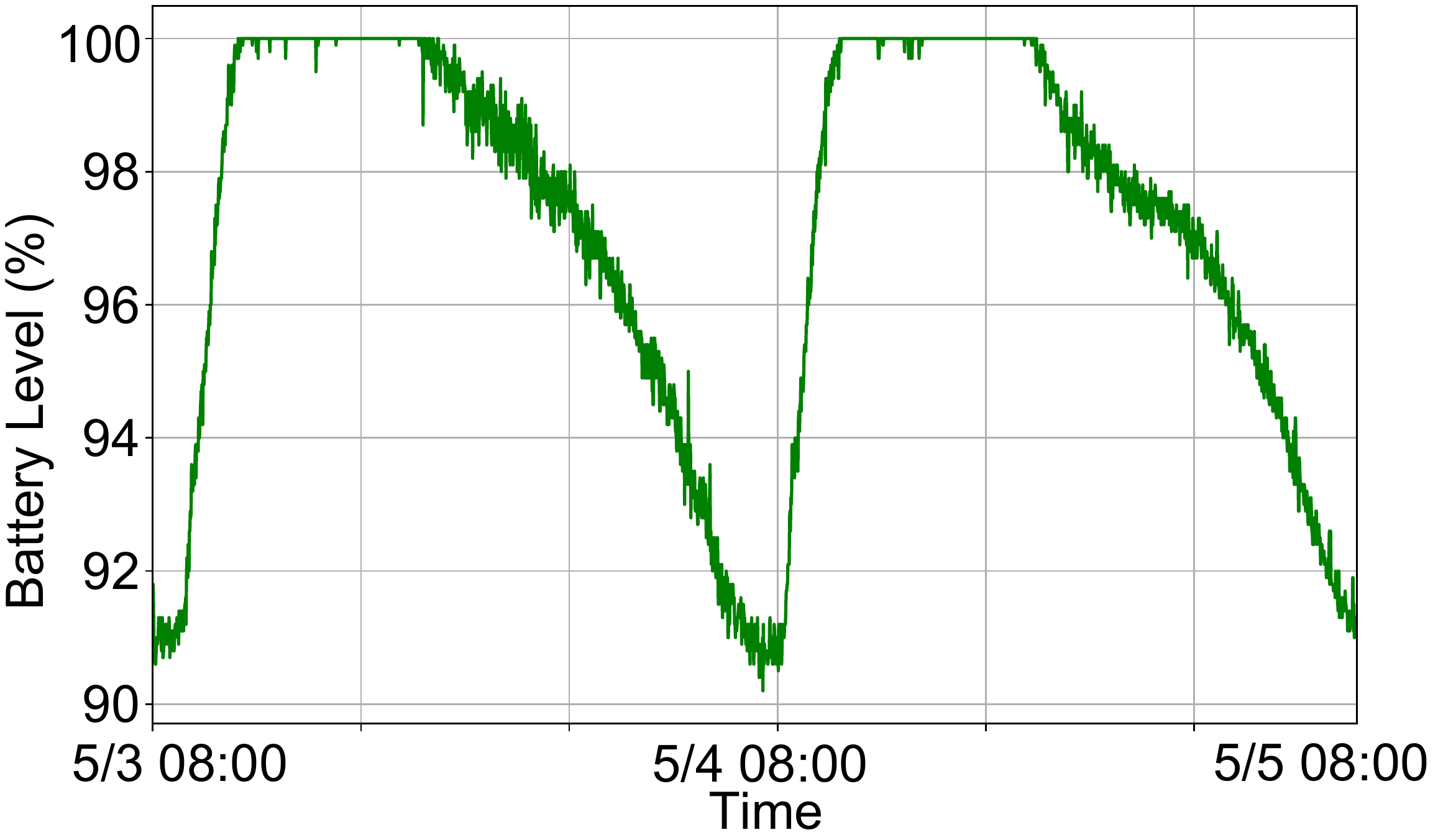}
		\caption{Battery Charging and Discharging Cycle.}
		\label{energy}
		
	\end{minipage}	

\end{figure*}

\subsection{Effect of proposed Reward.}
In this section, we discuss the simulation results of \aliasAPP with different rewards for a whole growing season (March 1st to October 31st, 246 days).

In order to minimize water consumption while not affecting plant health, we consider three situations in the reward. 1) The soil water content ($V_{i}$) is higher than the FC ($V_{fc}$) level. 2) $V_{i}$ is between $V_{fc}$ and $V_{mad}$. 3) $V_{i}$ is lower than $V_{mad}$. Only in the second situation, the plants are in good health. 
To evaluate our reward function, we compare it with a simple reward ($\aliasAPP_{MAD}$) that only maintains $V_{i}$ above $V_{mad}$. 
It is commonly used in the sensor-based method \cite{grabow2013water}. The reward is defined as: $R = -\sum_{i=1}^{N}\lambda_{3}*( V_{mad}  - V_{i} )    +    \mu_{3}*a_{i},  
 V_{i}<V_{mad}$. This function gives more penalty to plant health when $V_{i}$ is lower than $V_{mad}$ since plants’ health is significantly impacted. All the parameters are the same in Section \ref{reward_section}.

\begin{figure}[t]
  \includegraphics[height=1.8in, width=3in]{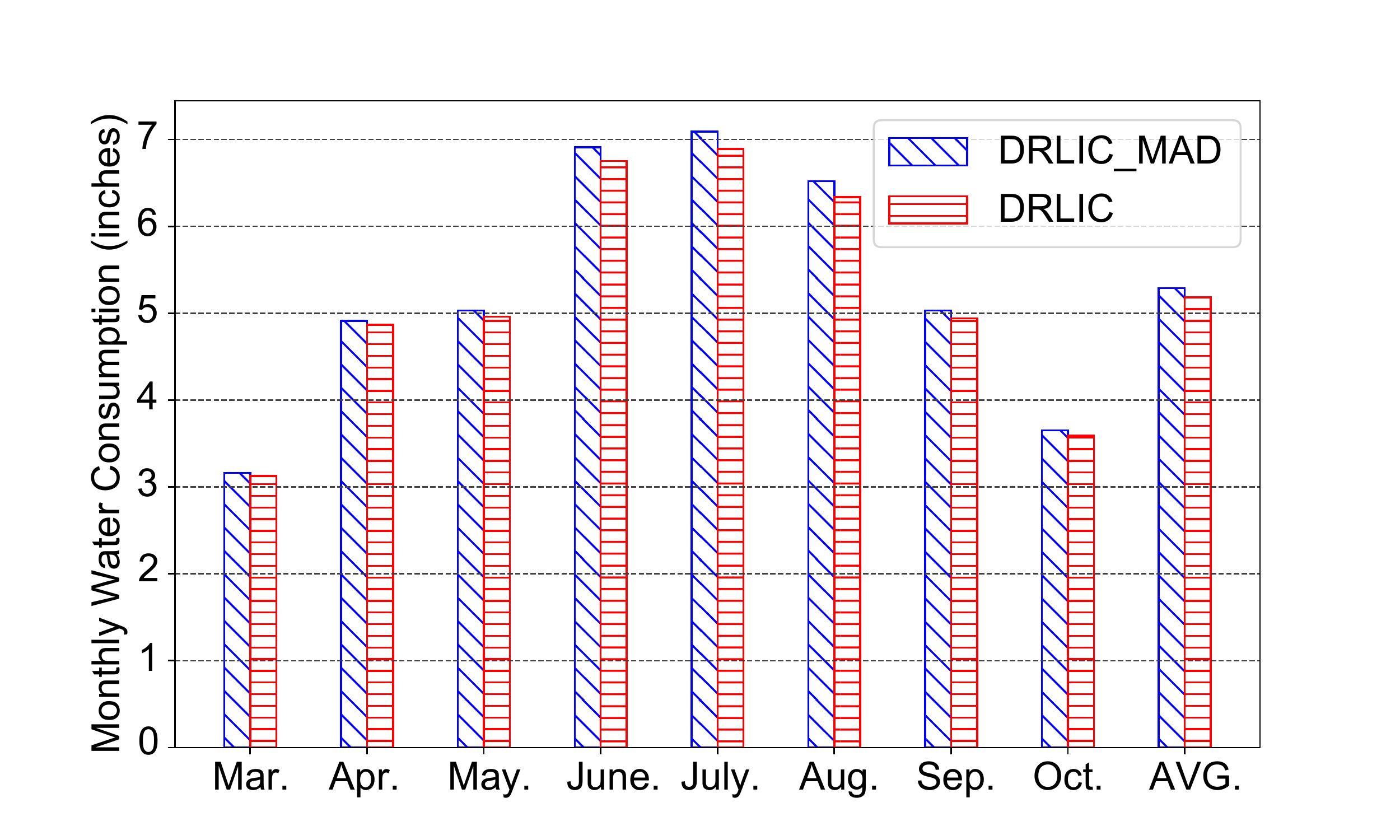}
    \vspace{-0.15in}
  \caption{Water Consumption for \aliasAPP with Different Reward}
  \label{reward_water_consumption}
\end{figure}

Figure \ref{reward_water_consumption} shows the water consumption of \aliasAPP with our proposed reward (\aliasAPP) and the simple reward (\aliasAPP\_MAD). 
\aliasAPP can save 2.04\% more water than \aliasAPP\_MAD, as the latter does not consider the case when $V_{i}$ is higher than $V_{mad}$. 
\aliasAPP considers two more situations by giving different penalties to plants’ health and water consumption. 
The first case is over-irrigation. The water consumption is too high. Therefore, the penalty for water consumption is higher than plant health. 
In the second case, the plants are in good health. \aliasAPP strives to maintain the $V_{i}$ close to $V_{mad}$ to save more water.

\subsection{\aliasAPP Policy Convergence.}
Figure \ref{reward_converage} shows the RL training process and the policy converges around the 500th training iteration. We define the length of an episode as 30 days.  We randomly vary the soil water content for each tree between the FC (7.08 inches) and MAD (4.72 inches) at the beginning of each episode. By doing
so, the policy is exposed to different soil water content conditions and learns to avoid water depletion than the MAD level during training. At the beginning of
the experiment, the RL policy receives a larger negative reward as it does
not know a valid sequence of actions that maximize the reward. The policy converges at the 500th training iteration.  The whole training
(i.e. 1000 training iterations) takes $\sim $ 4 hours using a 64-bit quad-core Intel Core i5-7400 CPU at 3.00 GHz.

\begin{table}[t]
  \renewcommand\arraystretch{0.8}
  \caption{ Micro-sprinkler Node Manufacture Cost.}
  \vspace{-0.1in}
  \centering
  \begin{tabular}{c|c||c|c}
    \hline
     \textbf{Component} & \textbf{Price}&\textbf{Component} & \textbf{Price}
\\
    \hline
    
    Moisture Sensor x 2 
  & \$250 &ESP32    & \$6.5\\

    18650 Li-ion battery    & \$3 &Solar Panel   & \$4.3 \\ 

    Latching Solenoid 
     &\$4 &Switch Relay x 2  & \$5\\
     
    Waterproof Enclosure  & \$12&Maintenance Fee& \$10\\
    \hline
      & & Total& \$294.8\\    
    

  \end{tabular}
  \label{node_cost}
\end{table}
\subsection{Energy Consumption of Sensor Nodes}
From a wireless sensor network standpoint, the ability of a system to operate for a long period of time without user intervention is fundamental. \aliasAPP nodes are no different, especially if they are meant to be put on the ground. For this reason, our hardware and software were designed to consume as little energy as possible. \aliasAPP nodes were fitted with a latching solenoid, allowing the flow of water to be turned on or off with a short pulse of power, rather than a constant supply.  For additional energy savings, the radio in each node is duty-cycled, activating for only a 10 second period every 1 minute. We need this high data frequency, the reason is that the base station can send an off command to \aliasAPP with a minute granularity. In our devices, the four peripherals that consume significant energy are the two moisture sensors, solenoid, two relays and radio. To meet this energy, we design an energy harvesting mechanism by leveraging one 5/6 V 1.2 W solar panel.

Figure \ref{energy_parts} shows the energy consumption for different sensors. Each moisture sensor sample requires 10 mA of power for 10 ms, and each flip of the latching solenoid requires 380 mA of power for 30ms. The ESP32 radio requires 180 mA of power for 50ms when in transmitting mode. The relay requires 250 mA for 20 ms for switching on or off. In our system, to ensure we don’t cut power too early, we add a safety band of 50\% on the timing on both of these devices, triggering for 15 ms and 45 ms for the sensor and solenoid, respectively. Overall, the solar-harvest mechanism can meet the daily requirement of all the sensors in \aliasAPP node.

Figure \ref{energy} shows the two days' energy charging and discharging process. After a night discharge, the 18650 battery level is increasing at 9:15 am on May 3rd. It usually takes 2 hours to fully charge the battery (9:15 - 11:35 am). The battery level will keep 100\% from 11:35 am to 18:45 pm, the energy harvested from solar can meet the energy requirement of all sensors in \aliasAPP node. The battery will discharge from 100\% at 18:45 pm of May 3rd to the 90.7\% at 8:45 am of May 4th. Then the whole energy charging and discharging process repeat. The lowest battery level is an average of 90\%. In the 2 week's deployment, we find that even on a cloudy day, the battery can also be charged and will take one more hour to be fully charged.

\subsection{Return on Investment}
 \label{return}

A primary concern to purchasing or upgrading an irrigation control system is the return on investment, i.e., how long does it take to save enough money from water consumption to cover the cost of the new irrigation system. 
To calculate the return on investment of \aliasAPP, we take into account the initial investment cost of the \aliasAPP system and the money saved from the less water consumption provided by our increased irrigation efficiency.

We first calculate the cost to develop a single \aliasAPP node. 
All the components of a \aliasAPP node can be found in a consumer electronics store and a home improvement store. 
Table \ref{node_cost} lists the cost of all components. In total, a \aliasAPP sensing and actuation node costs \$294.8. A large portion of the budget is the cost of two soil moisture sensors. We use two expensive soil moisture sensors that provide accurate measurement and a long lifetime.

The factors that mostly influence the payback of our system are water price and water volume saved by \aliasAPP. Water price varies considerably in different irrigation district and over time. 
This study assumed 100\% groundwater usage and availability. 
Each tree costs \$11.3 for irrigation water per month. Based on our experiment results, \aliasAPP can save 9.52\% of water expense per month, corresponding to \$1.08. Normally, almond orchards have 100 trees per acre. As a result, \aliasAPP can save \$108 per month.
Take a 60-acre almond orchard with 10 irrigation regions as an example.
Each irrigation region is six acres.
\aliasAPP can save \$648 in each irrigation region per month.

In each irrigation region, we need to deploy one \aliasAPP node, which costs \$294.8. 
The other irrigation components will use the existing infrastructure, such as the pipelines and micro-sprinklers under each tree. 
The cost of upgrading the existing irrigation system with our irrigation control system is \$294.8 for one irrigation region in an orchard. 
Every month, our system can save \$648. 
Therefore, it only needs half a month for our irrigation system to return the investment. 





\section{Related work}
\label{relate_work}


\textbf{ET-Based Irrigation Control.} 
As the weather is a primary water source or sinks in an irrigated space, systems have been developed to use weather as input for control. The simplest of these systems use standard fixed-schedule irrigation, but allow a precipitation sensor to override control to save water during rain \cite{Hunter}. The more complicated systems, now the industry standard, use evapo-transpiration (ET), an estimate of the amount
of water lost to evaporation and plant transpiration to do efficient
water-loss replacement \cite{allen1998crop, jensen1990evapotranspiration}. Some providers boast an average 30\% reduction in water consumption, but as with all industry irrigation systems, ET-based systems are limited by centralized control, and can not provide site-specific irrigation, reducing potential system efficiency and quality of control. 


\textbf{Sensor-based Irrigation Control.}
With the introduction of more accurate and efficient soil moisture
sensors, work has been done to create irrigation controllers
that react directly to moisture levels in the soil \cite{grabow2013water, UGMO, kim2008remote}. Moisture sensors buried in the root zone of trees accurately measure the moisture level in the soil and transmit this data to the controller. The controller then adjusts the pre-programmed watering schedule as needed. There are two types of soil moisture sensor-based systems: 1) Suspended cycle irrigation systems. Suspended cycle irrigation systems use traditional timed controllers and automated watering schedules, with start times and duration. The difference is that the system will stop the next scheduled irrigation cycle when there is enough moisture in the soil. 2) Water on-demand irrigation requires no programming of irrigation duration (only start times to water). This type maintains two soil moisture thresholds. The lower one to initiate watering, and the upper one to terminate watering \cite{grabow2013water}. However, without a model of the way water is lost, these thresholds are usually set based on experience and are not optimal.

\textbf{Model-based Irrigation Control.}
In \cite{winkler2016magic, winkler2019dictum}, a mechanistic PDE model of moisture movement within irrigated space is built. Using this model, an optimal watering schedule can be found to maintain a proper moisture level. However, the PDE model is not updated over time and future weather prediction is not taken into account. To tackle these two limitations, the same authors further improve the control system in \cite{winkler2018plug, winkler2020optics}.
The PDE model is eschewed in favor of an adaptive approach that involves models trained from sensor data. Long-term and short-term models are developed to describe the relationship of runoff between sprinklers in the movement of water through the soil.

As indicated by the authors \cite{winkler2018plug, winkler2020optics}, their system is designed for turf irrigation, and it is unlikely to provide benefit in shrubbery or tree irrigation. First, the turf soil moisture is affected by water runoff on soil surface and the overlapping coverage of sprinklers. The models in \cite{winkler2018plug, winkler2020optics} are focused on capturing the relationship of runoff between sprinklers. For tree irrigation, however, there is little runoff due to the tree space. 
The soil moisture model for tree irrigation needs to consider the soil-water relationship under different depths. 
Second, as shown in \cite{murthy2019machine}, the decay of volumetric water content derived from the long-term model of \cite{winkler2018plug} was shown to be much quicker than the real-world scenarios. It is bound to irrigate lightly and frequently, which has been found to be inefficient \cite{LF}.

\textbf{DRL-based Control.}
DRL has been applied in many applications, such as network planning \cite{zhu2021network}, cellular data analytics \cite{shen2020dmm}, sensor energy management \cite{fraternali2020ember}, mobile app prediction \cite{shen2021deepapp, yang2021atpp} and building energy optimization \cite{ding2023exploring, ding2023multi}. In particular, DRL techniques have demonstrated the potential optimal irrigation controls. \cite{sun2017reinforcement} proposed a reinforcement learning-based irrigation control system. The basic idea is to use a reinforcement learning algorithm to perform both irrigation planning and scheduling. Two neural
networks (NNs) are also introduced to predict DSSAT (Decision Support System for Agrotechnology Transfer) \cite{jones2003dssat} simulation results. DSSAT is the defacto standard model for crop growth. One NN inputs irrigation and weather information and predicts total soil water content, while the other NN predicts crop yield, given the daily total soil water content for an entire crop season. The prediction of crop yield is then used as the training data to train the reinforcement learning model. This approach can achieve relatively precise irrigation and allows full automation of the irrigation process. However, it is restricted to a small state space and is difficult to scale to large problems. Therefore, accurate representation of the actual irrigation context is difficult, leading to loss of important information that is needed for optimizing irrigation decisions. To solve this small state space problem, a deep reinforcement learning-based irrigation scheduling approach \cite{yang2020deep} is introduced for optimizing irrigation applications in terms of net return. This approach determines the amount of irrigation for each zone at each time step, taking soil moisture, evapotranspiration, precipitation probability, and crop growth stage into consideration. Compared to the previous approach using traditional reinforcement learning, it can handle a much greater state space and a greater number of irrigation choices. However, this control method is still central-valve control and thus affects the production performance.

\textbf{TestBed} TestBeds are widely used to study the precision irrigation. There are several existing platforms with features required for irrigation control \cite{farmbot, momin2019design, wiggert2019rapid, presten2022design}. The Farmbot is a CNC style mechanism that consists of a plot for vegetables and a modular set of interchangeable tools \cite{farmbot, momin2019design}. The tools can execute a variety of tasks including soil moisture sensing, RGB
imaging, planting, weeding, and irrigating. The basic assembly kit is priced at USD
\$2595. \cite{wiggert2019rapid, presten2022design} presented RAPIDMOLT, a modular, open-source testbed that enables real-time, fine-grained data collection and irrigation actuation. RAPIDMOLT costs USD \$600 and has floor space of 0.37 $m^2$. The functionality of the platform is evaluated by measuring the correlation between plant growth (Leaf Area Index) and water stress (Crop Water Stress Index) with irrigation volume. Both these two TestBeds are used for vegetables, not for trees with much deep soil and large scale.




\vspace{-0.05in}
\section{Conclusions}\label{sec:conclusion}
We present \aliasAPP, a DRL-based irrigation system that generates optimal irrigation control commands according to current soil water content, current weather data and forecasted weather information. A set of techniques have been developed, including our customized
design of DRL states and reward for optimal irrigation, a validated soil moisture simulator for fast DRL training, and a safe irrigation module. We design \aliasAPP irrigation node and build a testbed of six almond trees. Extensive experiments in real-world and simulation show the efficiency of \aliasAPP system.

\section{Acknowledgments}\label{sec:acknowledgments}

We would like to thank our anonymous shepherd and reviewers for their constructive comments. We also thank Danny Royer for helping us set up the testbed. This research is partially supported by the National Science Foundation under grants \#CCF- 2008837, a 2020 Seed Fund Award from CITRIS and the Banatao Institute at the University of California, and a 2022 Faculty Research Award through the Academic Senate Faculty Research Program at the University of California, Merced.

\balance
\bibliographystyle{unsrt}
\bibliography{acm}

\end{document}